%% file: arxiv_main.tex
\documentclass[10pt,twocolumn,letterpaper]{article}

\usepackage{iccv}
\usepackage{times}
\usepackage{epsfig}
\usepackage{graphicx}
\usepackage{amsmath}
\usepackage{amssymb}
\usepackage[accsupp]{axessibility}

\usepackage{booktabs}

\usepackage[breaklinks=true,bookmarks=false]{hyperref}

\iccvfinalcopy 

\def\iccvPaperID{6211} 

\ificcvfinal\fi

\begin{document}

\title{NeuS2: Fast Learning of Neural Implicit Surfaces for Multi-view Reconstruction}


\author{Yiming Wang*\\
\normalsize University of Pennsylvania, Peking University\\
{\tt\small wym12416@pku.edu.cn} 
\and
Qin Han*\\
\normalsize Peking University\\
{\tt\small hanqin@pku.edu.cn}
\and
Marc Habermann\\
\normalsize Max Planck Institute for Informatics\\
{\tt\small mhaberma@mpi-inf.mpg.de}
\and
Kostas Daniilidis\\
\normalsize University of Pennsylvania\\
{\tt\small kostas@cis.upenn.edu}
\and
Christian Theobalt\\
\normalsize Max Planck Institute for Informatics\\
{\tt\small theobalt@mpi-inf.mpg.de}
\and
Lingjie Liu\\
\normalsize University of Pennsylvania, Max Planck Institute for Informatics\\
{\tt\small lingjie.liu@seas.upenn.edu}
}

\maketitle

\let\svthefootnote\thefootnote
\newcommand\freefootnote[1]{%
  \let\thefootnote\relax%
  \footnotetext{#1}%
  \let\thefootnote\svthefootnote%
}

\freefootnote{$*$ Equal contribution}

\ificcvfinal\thispagestyle{empty}\fi

\input{Sec/0_Abstract}


\input{Fig/0_fig_teaser_tex}


\input{Sec/1_Intro.tex}
\input{Sec/2_Related_Work_new_reduced}
\input{Sec/3_Method_new_reduced}
\input{Sec/4_Experiments.tex}

\input{Sec/5_Conclusion}
\input{Sec/6_Acknowledgement}

{\small
\bibliographystyle{ieee_fullname}
\bibliography{egbib}
}

\newpage
\section*{Supplementary Material}

\appendix


\input{supp/0_intro}
\input{supp/1_derivation}

\input{supp/1.5_concurrent_works}
\input{supp/2_dataset}
\input{supp/3_additional_results}
\input{supp/4_implemental_details}

\end{document}

%% file: Sec/0_Abstract.tex
\begin{abstract}
Recent methods for neural surface representation and rendering, for example NeuS~\cite{wang2021neus}, have demonstrated the remarkably high-quality reconstruction of static scenes. 
However, the training of NeuS takes an extremely long time (8$~$hours), which makes it almost impossible to apply them to dynamic scenes with thousands of frames.
We propose a fast neural surface reconstruction approach, called \textit{NeuS2}, which achieves two orders of magnitude improvement in terms of acceleration without compromising reconstruction quality. 
To accelerate the training process, we parameterize a neural surface representation by multi-resolution hash encodings and present a novel lightweight calculation of second-order derivatives tailored to our networks to leverage CUDA parallelism, achieving a factor two speed up. 
%
%
To further stabilize and expedite training, a progressive learning strategy is proposed to optimize multi-resolution hash encodings from coarse to fine. 
We extend our method for fast training of dynamic scenes, with a proposed incremental training strategy and a novel global transformation prediction component, which allow our method to handle challenging long sequences with large movements and deformations. 
%
Our experiments on various datasets demonstrate that NeuS2 significantly outperforms the state-of-the-arts in both surface reconstruction accuracy and training speed for both static and dynamic scenes. 
    The code is available at our website: \url{https://vcai.mpi-inf.mpg.de/projects/NeuS2/}. 
%
%
\end{abstract}

%% file: Fig/0_fig_teaser_tex.tex
%
%
\begin{figure}
	\centering
	\includegraphics[width=1.0\linewidth]{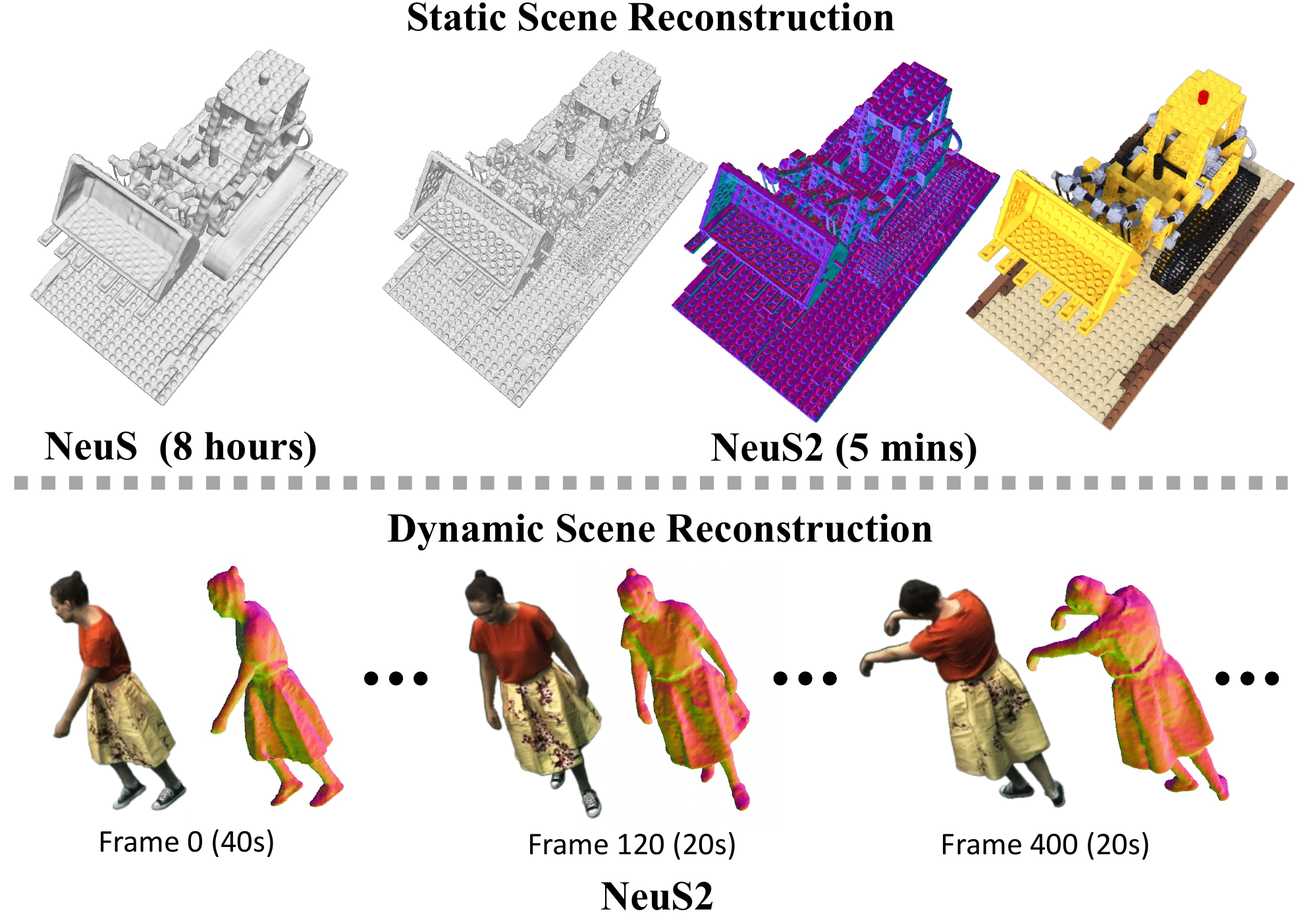}
	\vspace{-12pt}
	\caption
	{
	    We present NeuS2, a fast neural scene reconstruction method.
	    Given a set of multi-view images, NeuS2 can accurately reconstruct the scene geometry and appearance in the order of minutes.
	    This is in stark contrast to previous work~\cite{wang2021neus}, which only recovers medium-scale details at significantly increased time (about 8 hours).
	    Moreover, we demonstrate that NeuS2 can also be applied to dynamic scene reconstruction from multi-view videos, where we recover per-frame reconstruction in about 20 seconds.
	}
	\label{fig:teaser}
 \vspace{-10pt}
\end{figure}
%
%

%% file: Sec/1_Intro.tex
%
%
\vspace{-12pt}
\section{Introduction} \label{sec:intro}
%
%
Reconstructing the dynamic 3D world from 2D images is crucial for many Computer Vision and Graphics applications, such as AR/VR, 3D movies, games, telepresence, and 3D printing.
Classical stereo algorithms employ computer vision methods, e.g. feature matching, to capture the geometry and appearance of 3D contents from multi-view 2D images. 
Despite great progress, these methods are still comparably slow and struggle to reconstruct high-quality results. 
%
%
\par
Recently, 3D reconstruction with neural implicit representations has become a promising alternative to traditional methods, because of its high spatial resolution and highly detailed reconstructions outperforming classical stereo algorithms~\cite{barnes2009patchmatch,furukawa2009accurate,galliani2016gipuma, schonberger2016pixelwise,de1999poxels,broadhurst2001probabilistic, seitz1999photorealistic}. 
NeuS~\cite{wang2021neus}, as a representative work, models geometry surfaces as a neural network encoded Signed Distance Field (SDF), and renders an image via differentiable volume rendering. 
While NeuS~\cite{wang2021neus} produces high-quality reconstruction results, its training process is extremely slow, i.e. about 8 hours for a static object. 
This makes it nearly impossible to reconstruct dynamic scenes. 
Instant-NGP~\cite{muller2022instant_ngp} has explored the training acceleration of neural radiance fields (NeRF)~\cite{mildenhall2020nerf} by utilizing multi-resolution hash tables to augment neural network-encoded radiance fields. 
Though Instant-NGP~\cite{muller2022instant_ngp} synthesizes impressive novel view synthesis results, the extracted geometry from the learned density fields contains discernible noise since the 3D representation lacks surface constraints. 
%
%
\par
To overcome these drawbacks, we propose \textit{NeuS2}, a new method for fast training of highly-detailed neural implicit surfaces (see Fig.~\ref{fig:teaser}). 
\textit{NeuS2} reconstructs a static object in minutes, and a moving object sequence in up to 20 seconds per frame. 
To achieve this, we parameterize the neural network-encoded SDF using multi-resolution hash tables of learnable feature vectors~\cite{muller2022instant_ngp}.
%
Notably, in this design, a surface constraint and the rendering formulation require calculating the second-order derivatives. 
The main challenge is to have a simple and memory-efficient calculation 
to achieve the highest possible GPU computing performance.
%
Therefore, we derive a simple formula of the second-order derivatives tailored to ReLU-based MLPs, which enables an efficient CUDA implementation with a small memory footprint at a significantly reduced computational cost.  
To further enforce and accelerate the training convergence, we introduce an efficient progressive training strategy
, which updates the hash table features in a coarse-to-fine manner. 
%
%
%
\par
We further extend our method to multi-view dynamic scene reconstruction. 
Instead of training each frame in the sequence separately, we propose a new incremental learning strategy to efficiently learn a neural dynamic representation of objects with large movements and deformations. 
Specifically, we exploit the similarity of the shape and appearance information shared in two consecutive frames by first training the first frame and sequentially fine-tuning  the subsequent frames.  
%
%
While generally, this strategy works well, we observe that when the movement between two consecutive frames is relatively large, the predicted SDF of the occluded regions that are not observed in most images may get stuck in the learned SDF of the previous frame. 
To address this, we predict a global transformation to roughly align these two frames before learning the representation of the new frame. 
%
%
%
In summary, our technical contributions are:
\begin{itemize}
	\item We propose a new method, \textit{NeuS2}, for fast learning of neural surface representations from multi-view RGB input for both static and dynamic scenes, which achieves a significant speed up over the state-of-the-art while achieving an unprecedented reconstruction quality.
        \item A simple formulation of the second-order derivatives tailored to ReLU-based MLPs is presented to enable efficient parallelization of GPU computation. 
        \item A progressive training strategy for learning multi-resolution hash encodings from coarse to fine is proposed to enforce better and faster training convergence. 
        \item We design an incremental learning method with a novel global transformation prediction component for reconstructing long sequences (e.g., 2000 frames) with large movements in an efficient and stable manner.
\end{itemize}

%% file: Sec/2_Related_Work_new_reduced.tex
%
%
\section{Related Work} \label{sec:related}
%
%
\textbf{Multi-view Stereo.} 
Traditional multi-view 3D reconstruction methods can be categorized into depth-based and voxel-based methods. 
Depth-based methods~\cite{barnes2009patchmatch,furukawa2009accurate,galliani2016gipuma, schonberger2016pixelwise} reconstruct a point cloud by identifying point correspondences across images.
However, the reconstruction quality is heavily affected by the accuracy of correspondence matching.
Voxel-based methods~\cite{de1999poxels,broadhurst2001probabilistic, seitz1999photorealistic} side-step the difficulties in explicit correspondence matching by recovering occupancy and color in a voxel grid from multi-view images with a photometric consistency criterion. 
However, the reconstruction of these methods is limited to low resolution due to the high memory consumption of voxel grids. 
%
%
\par 
\textbf{Classical Multi-view 4D Reconstruction.} 
A large body of work~\cite{carranza2023,Vlasic2008,Ahmed2008,Aguiar2008,Xu2018,Habermann2019} in multi-view 4D reconstruction utilizes a precomputed deformable model, which is then fit to the multi-view images. 
In contrast, our method does not rely on a precomputed model, can reconstruct detailed results, and handles topology changes.
The most relevant to our work is \cite{Collet2015}, which is also a model-free method.
They leverage RGB and depth inputs to reconstruct high-quality point clouds for each frame, and then produce temporally coherent geometry. 
Instead, we only require RGB as input and can learn the high-quality geometry and appearance for each frame in an end-to-end manner in 20 seconds per frame. 
%
%
\par 
\textbf{Neural Implicit Representations.}
Neural implicit representations have made remarkable achievements in novel view synthesis~\cite{sitzmann2019scene,lombardi2019neural,kaza2019differentiable,mildenhall2020nerf,liu2020neural, sitzmann2019deepvoxels} and 3D/4D reconstruction~\cite{saito2019pifu,saito2020pifuhd,yariv2020multiview,niemeyer2020differentiable,kellnhofer2021neural,jiang2020sdfdiff,liu2020dist,Oechsle2021ICCV,wang2021neus,yariv2021volume}.
NeRF~\cite{mildenhall2020nerf} has shown high-quality results in the novel view synthesis task, but it cannot extract high-quality surfaces since the geometry representation lacks surface constraints. 
NeuS~\cite{wang2021neus} represents the 3D surface as an SDF for high-quality geometry reconstruction.
However, the training of NeuS is very slow, and it only works for static scenes. 
Instead, our method is 100 times faster and can be further accelerated to 20 seconds per frame when applied to dynamic scene reconstruction. 
\par 
Some NeRF-based works~\cite{yu_and_fridovichkeil2021plenoxels,SunSC22,muller2022instant_ngp} introduce voxel-grid features to represent 3D properties for fast training. 
However, these methods cannot extract high-quality surfaces as they inherit the volume density field as the geometry representation from NeRF~\cite{mildenhall2020nerf}. 
In contrast, our method can achieve high-quality surface reconstruction as well as fast training. 
For dynamic scene modeling, many works~\cite{tretschk2020nonrigid,park2020nerfies, pumarola2020dnerf, park2021hypernerf, Li2020NeuralSF, xian2020spacetime,li20223dvideo_synthesis} propose to disentangle a 4D scene into a shared canonical space and a deformable field per frame.

\cite{tineuvox} represents a 4D scene with a time-aware voxel feature. 
\cite{liu2022devrf} proposes a static-to-dynamic learning paradigm for fast dynamic scene learning.
\cite{li2022streaming_video} presents a grid based method for efficiently reconstructing radiance fields frame by frame. 
\cite{wang2022fourier} presents a novel Fourier PlenOctree method to compress a dynamic scene into one model.
These four methods focus on the novel view synthesis and, thus, are not designed to reconstruct high-quality surfaces, which is different from our goal of achieving high-quality surface geometry and appearance models. 
While these works make the training for dynamic scenes more efficient, the training is still time-consuming. 
Furthermore, these methods are not able to handle large movements and only reconstruct medium-quality surfaces. 
Some works in human performance modeling~\cite{su2021anerf, liu2021neural, peng2021animatable, chen2021animatable, xu2021hnerf, 2021narf, zhang2021stnerf, peng2020neural, kwon2021neural, Gao2022neuralnovelactor} can model large movements by introducing a deformable template as a prior.
In contrast, our method can handle large movements, does not require a deformable template and, thus, is not restricted to a specific dynamic object. 
Moreover, we can learn high-quality surfaces of dynamic scenes for 20 seconds per frame. 
\cite{zhao2022human} proposes a method for human modeling and rendering. 
It first reconstructs a neural surface representation for each frame; then it applies non-rigid deformation to obtain a temporally coherent mesh sequence.
Our work focuses on the first part, that is, fast reconstruction of dynamic scenes, where we exploit the temporal consistency between two consecutive frames to accelerate the learning of dynamic representation. Therefore, our work is orthogonal to~\cite{zhao2022human}, and can be integrated into~\cite{zhao2022human} as the first step.
\par
\textbf{Concurrent Work.}
Voxurf~\cite{Voxurf} proposes a voxel-based surface representation for fast multi-view 3D reconstruction. 
While it enables 20x speedup over the baseline (i.e., NeuS~\cite{wang2021neus}), our proposed method is over 3x faster than Voxurf and achieves better geometry quality when compared with Voxurf's results reported in their paper, 
as shown in the Suppl. document.
%
Neuralangelo~\cite{li2023neuralangelo} presents a novel method that leverages multi-resolution hash grids with numerical gradient computation for neural surface reconstruction.
It can achieve dense and high-fidelity geometry reconstruction results for large-scale scenes from multi-view images with multiple delicate designs while sacrificing its training cost, 100x slower when compared to ours.
Also, Voxurf and Neuralangelo are not designed for dynamic scene reconstruction. 
Last, Unbiased4d~\cite{johnson2023unbiased} proposes a monocular dynamic surface reconstruction approach by extending the NeuS formulation for bending rays.
In stark contrast to our approach, their focus lies on proving that unbiasedness also holds in the case of ray bending and the challenging monocular setting, and less on the highest possible quality at the fastest speeds.

%
%

%% file: Sec/3_Method_new_reduced.tex
\input{Fig/6_fig_smooth_ek_tex}
\input{Fig/1_fig_method_tex}
%
%
\section{Background} \label{sec:background}
\textbf{NeuS.} 
Given calibrated multi-view images of a static scene, NeuS~\cite{wang2021neus} implicitly represents the surface and appearance of a scene as a signed distance field $f (\mathbf{x}): \mathbb{R}^3 \to \mathbb{R}$ and a radiance field $c (\mathbf{x}, \mathbf{v}):\mathbb{R}^3 \times \mathbb{S}^2 \to \mathbb{R}^3$, where $\mathbf{x}$ denotes a 3D position and $\mathbf{v} \in \mathbb{S}^2$ is a viewing direction. 
The surface $\mathcal{S}$ of the object can be obtained by extracting the zero-level set of the SDF $\mathcal{S} = \{\mathbf{x}\in \mathbb{R}^3|f(\mathbf{x})=0\}$.
To render an object into an image, NeuS leverages   volume rendering. 
Specifically, for each pixel of an image, we sample $n$ points $\{\mathbf{\mathbf{p}(t_i)}=\mathbf{o}+t_i\mathbf{v}|i=0,1,\dots, n-1\}$ along its camera ray, where $\mathbf{o}$ is the center of the camera and $\mathbf{v}$ is the view direction.
By accumulating the SDF-based densities and colors of the sample points, we can compute the color $\hat{C}$ of the ray. 
As the rendering process is differentiable, NeuS can learn the signed distance field $f$ and the radiance field $c$ from the multi-view images. 
However, the training process is very slow, taking about 8 hours on a single GPU. 
%
%
\par 
\textbf{InstantNGP.} 
To overcome the slow training time of deep coordinate-based MLPs, which is also a main reason for the slow performance of NeuS, recently, Instant-NGP~\cite{muller2022instant_ngp} proposed a multi-resolution hash encoding and has proven its effectiveness.
Specifically, Instant-NGP assumes that the object to be reconstructed is bounded in multi-resolution voxel grids. 
The voxel grids at each resolution are mapped to a hash table with a fixed-size array of learnable feature vectors.
For a 3D position $\mathbf{x} \in \mathbb{R}^3$, it obtains a hash encoding at each level $h^i (\mathbf{x})\in \mathbb{R}^d$ (d is the dimension of a feature vector, $i=1, ..., L$) by interpolating the feature vectors assigned at the surrounding voxel grids at this level. 
The hash encodings at all $L$ levels are then concatenated to be the multi-resolution hash encoding $h(\mathbf{x}) =\{h^i(\mathbf{x})\}_{i=1}^{L}\in \mathbb{R}^{L\times d}$.
Besides the hash encoding, another key factor to the training acceleration is the CUDA implementation of the whole system, which makes use of GPU parallelism.
While the runtime is significantly improved, Instant-NGP still does not reach the quality of NeuS in terms of geometry reconstruction accuracy.
%
%
\par 

\textbf{Challenges.} 
Given the above discussion, one can ask whether a naive combination of NeuS~\cite{wang2021neus} and Instant-NGP~\cite{muller2022instant_ngp} can unite the best of the two worlds, i.e. high 3D surface reconstruction quality and efficient computation.
We highlight that \textit{it is far from being trivial to achieve training as fast as Instant-NGP}~\cite{muller2022instant_ngp} \textit{and, meanwhile, a reconstruction as high-quality as NeuS}~\cite{wang2021neus}.
Specifically, to ensure high-quality surface learning, the Eikonal constraint used in NeuS~\cite{wang2021neus} is indispensable, as shown in Fig.~\ref{fig:smooth_ek} and Suppl. materials;  and the key challenge of adding the Eikonal loss to CUDA-based MLPs (a key factor to fast training in Instant-NGP~\cite{muller2022instant_ngp}) is how to calculate the second-order derivatives efficiently for backpropagation.
Instant-NSR~\cite{zhao2022human} addresses this by approximating the second-order derivatives using finite differences, which suffer from precision problems and it can cause unstable training. 
Instead, we propose a simple, precise, and efficient formulation of second-order derivatives tailored to MLPs (Sec.~\ref{sec:derivatives}), which leads to fast and high-quality reconstruction.
The superiority of our approach over Instant-NSR is shown in Tab.~\ref{table:DTU} and Fig.~\ref{fig:static}.
\par
For dynamic scene reconstruction, there are two key challenges: 
how to exploit the temporal information for acceleration, and how to handle long sequences with large movements and deformations. 
To address them, we first propose an incremental training strategy to exploit the similarity in geometry and appearance information shared across two consecutive frames, which enabling faster convergence (Sec.~\ref{sec:incremental}). 
To handle large movements and deformations, we propose a novel Global Transformation Prediction component, which prevents the predicted SDF training from getting stuck in a local minimum.
Further, it can bound a dynamic sequence in a small volume to save memory and to improve reconstruction accuracy (Sec.~\ref{sec:transform}).
%
%
\section{Static Neural Surface Reconstruction} \label{sec:method_static}
We first present how our formulation can effectively learn the signed distance field of a \textit{static} scene from calibrated multi-view images (see Fig.~\ref{fig:overview} a)). 
To accelerate the training process, we first demonstrate how to incorporate multi-resolution hash encodings~\cite{muller2022instant_ngp} for representing the SDF of the scene, and how volume rendering can be applied to render the scene into an image (Sec.~\ref{sec:representation}). 
%
%
%
Next, we derive a simplified expression of second-order derivatives tailored to ReLU-based MLPs, which can be efficiently parallelized in custom CUDA kernels (Sec.~\ref{sec:derivatives}). 
Finally, we adopt a progressive training strategy for learning multi-resolution hash encodings, which leads to faster training convergence and better reconstruction quality (Sec.~\ref{sec:progessive}). 
%
%
\subsection{Volume Rendering of a Hash-encoded SDF} \label{sec:representation}
For each 3D position $\mathbf{x}$, we map it to its multi-resolution hash encodings $h_\Omega(\mathbf{x})$ with learnable hash table entries $\Omega$.
As $h_\Omega(\mathbf{x})$ is an informative encoding of spatial position, the MLPs for mapping $\mathbf{x}$ to its SDF $d$ and color $c$ can be very shallow, which results in more efficient rendering and training without compromising quality.
%
%
\par
\textbf{SDF Network.}
In more detail, our SDF network 
%
\begin{equation} \label{eqn:sdf_net}
(d, \mathbf{g}) = f_{\Theta}(\mathbf{e}),  \qquad 
\mathbf{e} = (\mathbf{x}, h_\Omega(\mathbf{x})).
\end{equation}
%
is a shallow MLP with weights $\Theta$, which takes the 3D position $\mathbf{x}$ along with its hash encoding $h_\Omega(\mathbf{x})$ as input and outputs the SDF value $d$ and a geometry feature vector $\mathbf{g} \in \mathbb{R}^{15}$.
Concatenating the position serves as a geometry initialization~\cite{atzmon2020sal} leading to a more stable learning of the geometry. 
%
%
\par
\textbf{Color Network.}
The normal of $\mathbf{x}$ can be computed as 
%
\begin{equation} \label{eqn:normal}
\mathbf{n}=\nabla_{\mathbf{x}}d.
\end{equation}
%
where $\nabla_{\mathbf{x}}d$ denotes the gradient of the SDF with respect to $\mathbf{x}$.
We then combine the normal with the geometry feature $\mathbf{g}$, the SDF $d$, the point $\mathbf{x}$, and the ray direction $\textbf{v}$ serving as the input to our color network
%
\begin{equation} \label{eqn:color_net}
\mathbf{c} = c_{\Upsilon}(\textbf{x}, \textbf{n}, \textbf{v}, d, \mathbf{g}),
\end{equation}
%
which predicts the color $\mathbf{c}$ of $\mathbf{x}$. 
%
%
\par
\textbf{Volume Rendering.}
To render an image, we apply the unbiased volume rendering of NeuS~\cite{wang2021neus}.
Additionally, we adopt a ray marching acceleration strategy used in Instant-NGP\cite{muller2022instant_ngp}. More details are provided
in the supp. document.
%

%
%
\par
\textbf{Supervision.}
To train NeuS2, we minimize the color difference between the rendered pixels $\hat{C_i}$ with $i \in \{1,...,m\}$ and the corresponding ground truth pixels ${C_i}$ without any 3D supervision. 
Here, $m$ denotes the batch size during training.
We also employ an Eikonal term~\cite{gropp2020igr} to regularize the learned signed distance field leading to our final loss
%
\begin{equation} \label{eqn:loss}
    \mathcal{L}=\mathcal{L}_{\mathrm{color}}+\beta \mathcal{L}_{\mathrm{eikonal}},
\end{equation}
%
where $\mathcal{L}_{\mathrm{color}} = \frac{1}{m}\sum_{i} \mathcal{R}(\hat{C_i}, C_i)$, and $\mathcal{R}$ is the Huber loss~\cite{huber}. $\mathcal{L}_{\mathrm{eikonal}} = \frac{1}{mn}\sum_{k,i}(||\mathbf{n}_{k,i}||-1)^2$, and $k$ indexes the $k$th sample along the ray with $k \in \{1,...,n\}$, and $n$ is the number of sampled points.
$\mathbf{n}_{k,i}$ is the normal of a sampled point (see Eq.~\ref{eqn:normal}). 
\par 
%
%
%
\subsection{Efficient Handling of Second-order Derivatives} \label{sec:derivatives}
To avoid the computational overhead that learning frameworks introduce, we implement our whole system in CUDA. In contrast to Instant-NGP~\cite{muller2022instant_ngp}, which only requires the first-order derivatives during the optimization, we must calculate the second-order derivatives for the parameters associated with the normal term $\mathbf{n}=\nabla_{\mathrm{x}}d$ (Eq.~\ref{eqn:normal}), which is input to the color network $c_\Upsilon$ (Eq.~\ref{eqn:color_net}) and the Eikonal loss term $\mathcal{L}_{\mathrm{eikonal}}$. 
%
%
%
%
\par
\textbf{Second-order Derivatives.}
To accelerate this computation, we directly calculate them using simplified formulas instead of applying the computation graph of PyTorch~\cite{NEURIPS2019_9015pytorch}. 
Specifically, we calculate the second-order derivatives of the hash table parameters $\Omega$ and the SDF network parameters $\Theta$ using the chain rule as
%
\begin{equation} \label{eqn:dloss_denc}
\begin{aligned}
\frac{\partial \mathcal{L}}{\partial \Omega} = \frac{\partial \mathcal{L}}{\partial \mathbf{n}} (
\frac{\partial \mathbf{e}}{\partial \mathbf{x}} \frac{\partial \frac{\partial d}{\partial \mathbf{e}}}{\partial \mathbf{e}} \frac{\partial \mathbf{e}}{\partial \Omega}
+ \frac{\partial d}{\partial \mathbf{e}} \frac{\partial \frac{\partial \mathbf{e}}{\partial \mathbf{x}}}{\partial \Omega}
)
\end{aligned}
\end{equation}
%
\vspace{-5pt}
%
\begin{equation} \label{eqn:dloss_dsdf}
\begin{aligned}
\frac{\partial \mathcal{L}}{\partial \Theta} 
&= 
\frac{\partial \mathcal{L}}{\partial \mathbf{n}} (
\frac{\partial \mathbf{e}}{\partial \mathbf{x}} \frac{\partial \frac{\partial d}{\partial \mathbf{e}}}{\partial \Theta} 
+ \frac{\partial d}{\partial \mathbf{e}} \frac{\partial \frac{\partial \mathbf{e}}{\partial \mathbf{x}}}{\partial \Theta}
)
\end{aligned}
\end{equation}
%

%
Note that the color network $c_\Upsilon$ only takes $\mathbf{n}$ as input, so we do not need to calculate its second-order gradients of the color network parameters $\Upsilon$. 
The derivation of Eq.~\ref{eqn:dloss_denc} and ~\ref{eqn:dloss_dsdf} is included in the supplementary document.
%
%
%
\par
To speed up the computation of Eqs.~\ref{eqn:dloss_denc} and \ref{eqn:dloss_dsdf}, we found that ReLU-based MLPs can greatly simplify the above terms leading to less computation overhead.
In the following, we discuss this in more detail and provide the proof of this proposition in the supplementary document. 
We first introduce some useful definitions as follows.
\newtheorem{theorem}{Theorem}
\newtheorem{definition}{Definition}
%

\begin{definition} \label{def:mlp}

Given a ReLU based MLP $f$ with $L$ hidden layers taking $x \in \mathbb{R}^{d}$ as input, it computes the output $y = H_L g(H_{L-1} \dots g(H_1x))$, where $H_l \in \mathbb{R}^{n_l} \times \mathbb{R}^{n_{l-1}}$, $l\in \{1,\dots,L\}$ is the layer index, and $g$ is the ReLU function. 
We define $P_l^j \in \mathbb{R}^{n_{l-1}} \times \mathbb{R}^1 $ and $S_l^i \in \mathbb{R}^ 1 \times \mathbb{R}^{n_{l}}$ as
%

\begin{equation}
\begin{aligned}
P_l^j &= G_{l} H_{l-1} \dots G_2 H_1^{(\_,j)} \\ 
S_l^i &= H_L^{(i,\_)} G_{L} \dots H_{l+1} G_{l+1}
\end{aligned}
\end{equation}
%
where $H_1^{(\_,j)}$ is the $j$th  column of $H_1$, $H_L^{(i, \_)}$ is the $ i$th row of $H_L$, and 
$G_l=
\begin{cases}
1, H_{l-1}\dots g(H_1x) > 0 \\
0, \text{ otherwise} 
\end{cases}$.
\end{definition}
Now the second order derivates of a ReLU-based MLP with respect to its input and intermediate layers can be defined.
%
\begin{theorem} [Second-order derivative of ReLU-based MLP] \label{theorem:mlp_derivative}
Given a ReLU based MLP $f$ with $L$ hidden layers with the same definition in Definition~\ref{def:mlp}. The second-order derivative of the MLP $f$ is:
\begin{equation} \label{eqn:mlp_2ord_derivative_formula}
\frac{\partial \frac{\partial y}{\partial x}_{(i,j)}}{\partial H_l} = (P_l^j S_l^i)^T
,\qquad\frac{\partial^2 y}{\partial \mathbf{x}^2} = 0
\end{equation}
where $\frac{\partial y}{\partial x}_{(i,j)}$ is the matrix element (i,j) of $\frac{\partial y}{\partial x}$, and $S_l^i$ and $P_l^j$ are defined in Definition \ref{def:mlp}.
\label{theo}
\end{theorem}
%
Coming back to our original second-order derivatives (Eqs.~\ref{eqn:dloss_denc} and \ref{eqn:dloss_dsdf}), by Theorem~\ref{theo}, we obtain $\frac{\partial \frac{\partial d}{\partial \mathbf{e}}}{\partial \mathbf{e}} = 0$. Since $\frac{\partial \mathbf{e}}{\partial \mathbf{x}}$ is irrelevant to $\Theta$, we have $\frac{\partial \frac{\partial \mathbf{e}}{\partial \mathbf{x}}}{\partial \Theta} = 0$. 
This results in the following simplified form
%
\begin{equation} \label{eqn:simplified_dloss_denc}
\begin{aligned}
\frac{\partial \mathcal{L}}{\partial \Omega}
&= \frac{\partial \mathcal{L}}{\partial \mathbf{n}} \frac{\partial d}{\partial \mathbf{e}} \frac{\partial \frac{\partial \mathbf{e}}{\partial \mathbf{x}}}{\partial \Omega}
\end{aligned}
\end{equation}
\vspace{-7pt}
\begin{equation} \label{eqn:simplified_dloss_dmlp}
\begin{aligned}
\frac{\partial \mathcal{L}}{\partial \Theta}
&= 
\frac{\partial \mathcal{L}}{\partial \mathbf{n}} 
\frac{\partial \mathbf{e}}{\partial \mathbf{x}} \frac{\partial \frac{\partial d}{\partial \mathbf{e}}}{\partial \Theta} 
\end{aligned}
\end{equation}
%
for the second-order derivatives, which result in improved efficiency and less computational overhead. 
The simplest form of $\frac{\partial \mathcal{L}}{\partial \Theta}$ can be obtained by substituting  $\frac{\partial \frac{\partial d}{\partial \mathbf{e}}}{\partial \Theta}$  using Eq.~\ref{eqn:mlp_2ord_derivative_formula}.
We show in Fig.~\ref{fig:ablation_2order_derivative} that our computation using Eq.~\ref{eqn:mlp_2ord_derivative_formula} is more efficient than PyTorch.
%
%
\subsection{Progressive Training} \label{sec:progessive}
Although our highly-optimized gradient calculation already improved training time, we found there is still room for improvement in terms of training convergence and speed.
Additionally, we empirically observed that using grids at low resolutions results in underfitting of the data, in which the geometry reconstruction is smooth and lacks details, whereas using grids at high resolutions induces over-fitting, leading to increased noise and artifacts of the results.
Therefore, we introduce a progressive training method by gradually increasing the bandwidth of our spatial grid encoding denoted as:
%
\begin{equation}
   h_\Omega(\mathbf{x}, \lambda) = \big(w_1(\lambda) h_\Omega^1(\mathbf{x}), \dots, w_{L}(\lambda)  h_\Omega^L(\mathbf{x}) \big), 
\end{equation}
%
where $h_\Omega^i$ is the hash encoding at level $i$ and the weight $w_i$ for each grid encoding level is defined by
%
$w_i(\lambda)=
I[i \leq \lambda]$,
%
and the parameter $\lambda$ modulates the bandwidth of the low-pass filter applied to the multi-resolution hash encodings.
%
Smaller parameter $\lambda$ leads to faster training speed, but limits the model’s capacity to
model high-frequency details. Thus, we initialize $\lambda$ as 2, and then gradually increase by 1 for every 2.5\% of the total training steps in all experiments. 
%
%
%
%
%
\section{Dynamic Neural Surface Reconstruction} \label{sec:method_dynamic}
We have explained how NeuS2 can produce highly accurate and fast reconstructions of static scenes. 
Next, we extend NeuS2 to dynamic scene reconstruction. 
That is, given multi-view videos of a moving object and camera parameters of each view, our goal is to learn the neural implicit surfaces of the object in each video frame (see Fig.~\ref{fig:overview} b)). 
%
%
\subsection{Incremental Training} \label{sec:incremental}
Even though our reconstruction method of static objects can achieve promising efficiency and quality, constructing dynamic scenes by training every single frame independently is still time-consuming.
However, scene changes from one frame to the other are typically small.
Thus, we propose an incremental training strategy to exploit the similarity in geometry and appearance information shared between two consecutive frames, which enables faster convergence of our model.
In detail, we train the first frame from scratch as presented in our static scene reconstruction, and then fine-tune the model parameters for the subsequent frames from the learned hash grid representation of the preceding frame.
Using this strategy, the model is able to produce a good initialization of the neural representation of the target frame and, thus, significantly accelerates its convergence speed.
%
%
\subsection{Global Transformation Prediction} \label{sec:transform}
%
As we observed during the incremental training process, 
the predicted SDF easily get stuck in the local minima of the learned SDF of the previous frame, especially when the object's movement between adjacent frames is relatively large.
For instance, when our model reconstructs a walking sequence from muli-view images, the reconstructed surface appears to have many holes, as shown in Fig.  \ref{fig:ablation_dynamic}. 
%
To address this issue, we propose a global transformation prediction to roughly transform the target SDF into a canonical space before the incremental training.
Specifically, we predict the rotation $R$ and transition $T$ of the object between two adjacent frames. 
For any given 3D position $\mathbf{x}_i$ in the coordinate space of the frame $i$, it is transformed back to the coordinate space of the previous frame $i-1$, denoted as $\mathbf{x}_{i-1}$
%
\begin{equation} \label{eqn:globalmove_predict}
    \mathbf{x}_{i-1}=R_i(\mathbf{x}_{i}+T_i).
\end{equation}
%
The transformations can then be accumulated to transform the point $\mathbf{x}_i$ back to $\mathbf{x}_c$ in the first frame's coordinate space
%
\begin{equation}
    \mathbf{x}_c=R_{i-1}^c(\mathbf{x}_{i-1}+T_{i-1}^c)=R_{i}^c(\mathbf{x}_{i}+T_{i}^c),
\end{equation}
%
where $R_i^c=R_{i-1}^cR_i$ and $T_i^c=T_i+R_i^{-1}T_{i-1}^c$.
%
%
\par
%
The global transformation prediction also allows us to model dynamic sequences with large movement in a small region, rather than covering the entire scene with a large hash grid. 
Since the hash grid only needs to model a small portion of the entire scene, we can obtain more accurate reconstructions and reduce memory cost.

%
%
\par
Notably, our method can handle large movements and deformations which are challenging for existing dynamic scene reconstruction approaches\cite{tretschk2020nonrigid}, \cite{pumarola2020dnerf}, thanks to the following designs of our approach: 
(1) Global transformation prediction that accounts for large global movements in the sequence; 
(2) Incremental training that learns relatively small deformable movements between two adjacent frames rather than learns relatively large movements from each frame to a common canonical space.  
%
%
\par
We realize incremental training combined with global transformation prediction as an end-to-end learning scheme, as illustrated in Fig.~\ref{fig:overview}(b).
When processing a new frame, we first predict the global transformation independently, and then fine-tune the model's parameters and global transformation together to efficiently learn the neural representation.

%% file: Fig/6_fig_smooth_ek_tex.tex
%
%
\begin{figure}
	\centering
	\vspace{-10pt}
	\includegraphics[width=\linewidth]{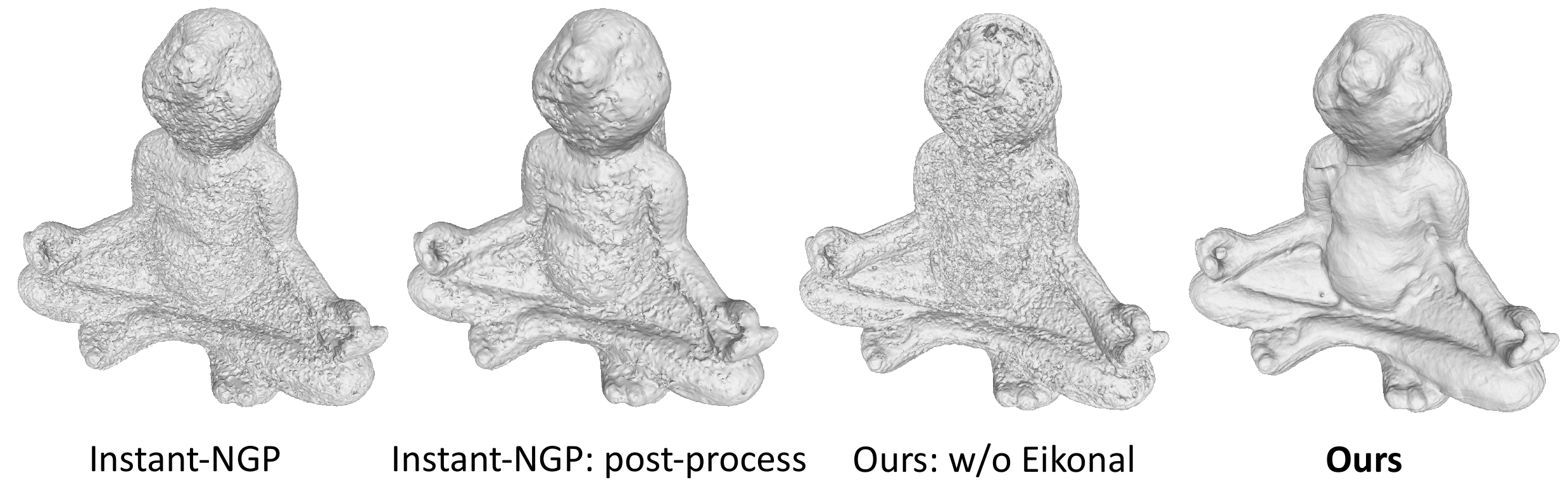}
	\caption
	{
	    %
            Qualitative comparison on DTU Scan 110, concerning the quality impact of smoothing post-process on Instant-NGP and the Eikonal Penalty on NeuS2.
	}
	\label{fig:smooth_ek}
	\vspace{-16pt}
\end{figure}
%
%

%% file: Fig/1_fig_method_tex.tex
%
%
\begin{figure*}
    \centering
	\includegraphics[width=0.8\linewidth]{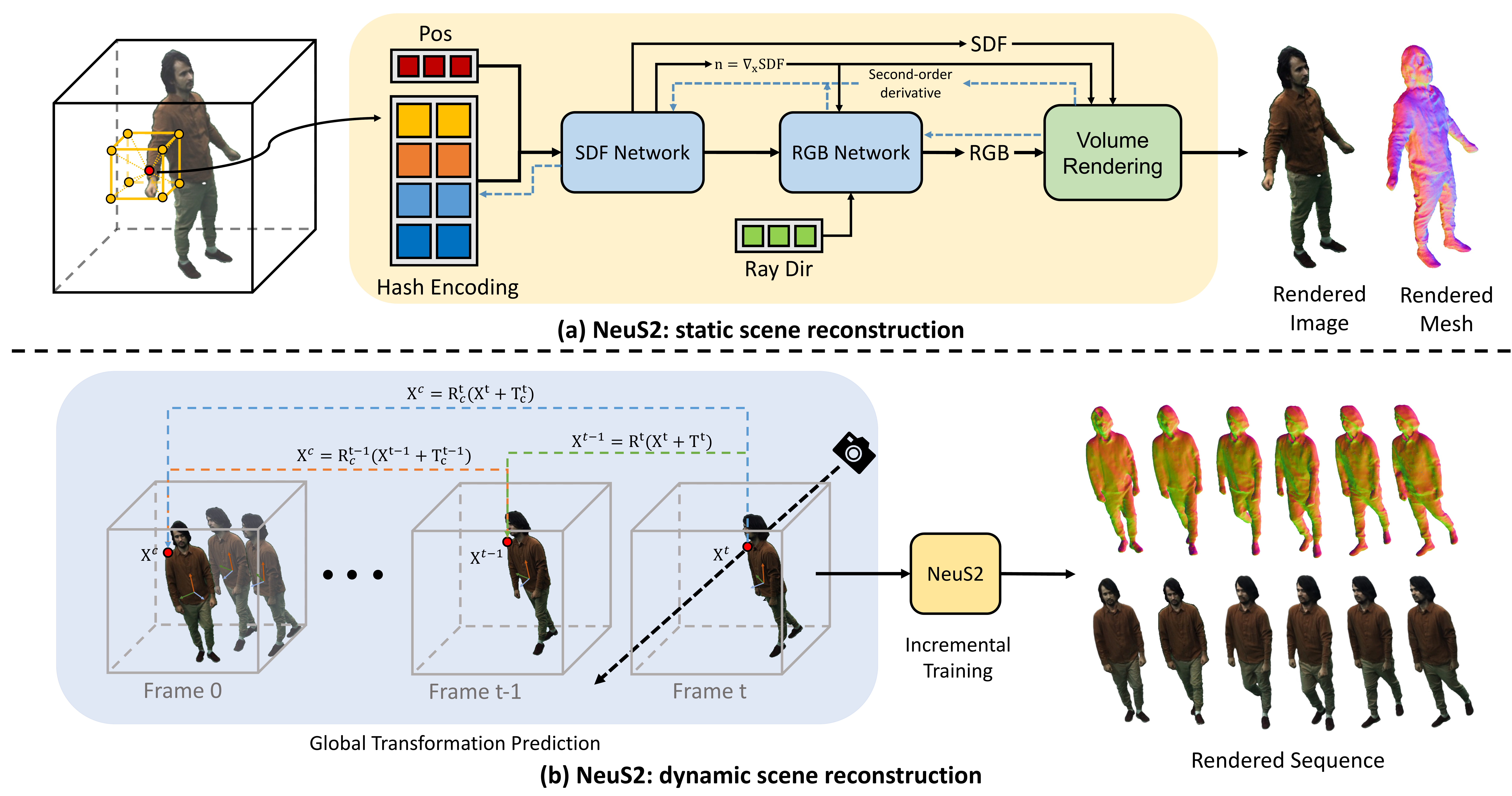}
	\vspace{-3pt}
	\caption
	{
	    Method overview.
	    %
	    (a) \textbf{Static Scene Reconstruction}: 
	    Given a 3D point $\mathbf{x}$, we concatenate its queried feature from the multi-resolution hash grid and its 3D position as the input to the SDF network.
	    The SDF network outputs the SDF value and geometry features, which are combined with the viewing direction and further fed into our RGB network to generate the RGB value. 
	    Notably, during backward propagation the second-order derivates are efficiently computed using our CUDA implementation.
	    (b) \textbf{Dynamic Scene Reconstruction}: 
	    Given a sequence of multi-view images, we first construct the first frame following our static reconstruction method. 
	    For every subsequent frame, we predict its global transformation with respect to the previous frame and accumulate the transformation to convert it into the canonical space (i.e. the first frame). 
	    Then, we fine-tune the parameters of NeuS2 for incremental training to generate the rendered results.
	}
	\label{fig:overview}
    \vspace{-10pt}
\end{figure*}
%
%

%% file: Sec/4_Experiments.tex
\section{Experiments} \label{sec:experiments}
All experiments are conducted on a single GeForce RTX3090 GPU.
Implementation details, additional results, and video results are provided in the supplementary material.
%
\input{Tab/0_tab_static}
\input{Fig/2_fig_static_tex}
\input{Tab/1_tab_dynamic_synthetic}
\input{Fig/3_fig_dynamic_synthetic_tex}
\input{Tab/2_tab_dynamic_real}
\input{Fig/4_fig_dynamic_real_tex}
%
%
\subsection{Static Scene Reconstruction}\label{sec:results_static}
For the static scene reconstruction, we use 15 scenes of DTU dataset~\cite{DTUdataset} for evaluation.
There are 49 or 64 images with a resolution of 1600 $\times$ 1200, and we test each scene with foreground masks provided by IDR~\cite{idr_yariv2020multiview}.  
To demonstrate the performance of \textit{NeuS2} on the static scene reconstruction task, we compare our approach with the-state-of-art methods NeuS~\cite{wang2021neus}, Instant-NGP~\cite{muller2022instant_ngp}, and Instant-NSR~\cite{zhao2022human}.  
We also provide a comparison with a concurrent work, Voxurf~\cite{Voxurf}, in the supplemental material.
\par 
For quantitative measurement, we use the Chamfer Distance~\cite{DTUdataset} to evaluate the geometry reconstruction quality, in the same way as NeuS~\cite{wang2021neus} did. 
We also measure the novel view synthesis quality by the peak signal-to-noise ratio (PSNR) between the reference images and the synthesized images. 
The quantitative comparison results are reported in Tab.~\ref{table:DTU}, and for per-scene breakdown results please refer to the Suppl. materials.
The results show that our method outperforms the baseline methods on the geometry reconstruction task with significantly less training time compared to NeuS~\cite{wang2021neus}.
Meanwhile, our method achieves comparable performance in the novel view synthesis task to Instant-NGP~\cite{muller2022instant_ngp} requiring the same training time (5 minutes).
\par 
A qualitative comparison of geometry reconstruction and novel view synthesis results for all methods is presented in Fig. \ref{fig:static}.
As shown in Fig. \ref{fig:static} for the 3D geometry reconstruction results, NeuS~\cite{wang2021neus} exhibits limited performance in terms of reconstructed details with excessively smooth surfaces.
The extracted meshes of Instant-NGP~\cite{muller2022instant_ngp}'s results are noisy since it lacks surface constraints in the geometry representation -- volume density field.
The results obtained from Instant-NSR~\cite{zhao2022human} exhibit many artifacts, which can be attributed to the use of a finite difference method to approximate the second derivative. This method is prone to precision problems and may cause unstable training. Furthermore, since Instant-NSR requires multiple forward calculations using the finite difference method, their approach is slower than ours.
Regarding the novel view synthesis results, our method outperforms NeuS~\cite{wang2021neus} and Instant-NSR~\cite{zhao2022human}, presenting detailed rendering results on par with Instant-NGP. 
To conclude, our method achieves high-quality geometry and appearance reconstruction, with additional details demonstrated in, both, the surface and rendered images without inducing noise, e.g. it can recover the complex structures of the windows and render detailed textures in scan 24.
%
%
\subsection{Dynamic Scene Reconstruction} \label{sec:results_dynamic}
We conduct experiments on both synthetic and real \textit{dynamic} scenes for the tasks of novel view synthesis and geometry reconstruction.
We compare our method with the state-of-the-art neural-based method for dynamic scenes, D-NeRF~\cite{pumarola2021d_nerf} and TiNeuVox~\cite{tineuvox} quantitatively and qualitatively. 
D-NeRF~\cite{pumarola2021d_nerf} and TiNeuVox~\cite{tineuvox} models general scenes for dynamic novel view synthesis by combining NeRF~\cite{mildenhall2020nerf} with deformation fields, which is learned from all the frames simultaneously.
Compared with D-NeRF which purely uses MLPs, TiNeuVox further accelerates the training by utilizing explicit voxel grids.
We also provide the results of Instant-NGP~\cite{muller2022instant_ngp} in the Suppl. materials.
\par 
\textbf{Synthetic Scenes.} 
We choose three different types of datasets for synthetic scene reconstruction, including a Lego scene shared by NeRF~\cite{mildenhall2020nerf} (150 frames), a Lion sequence provided by Artemis~\cite{luo2022artemis} (177 frames), and a human character in the RenderPeople~\cite{Renderpeople} dataset (100 frames).
For the quantitative evaluation, the Chamfer Distances and PSNR scores are calculated and averaged over all frames and all testing views of all frames, respectively.
As shown in Tab.~\ref{table:synthetics}, our method shows significantly improved novel view synthesis and geometry reconstruction results compared to D-NeRF and TiNeuVox.
Notably, our method takes less than 1 hour to complete the learning of a sequence, with 40 seconds (80 seconds for the Lego sequence) of training time for the first frame and 20 seconds for each subsequent frame; while the training time for each scene of D-NeRF is about 20 hours. 
The qualitative results are provided in Fig.~\ref{fig:dynamic_synthetic}, showing that our method outperforms D-NeRF and TiNeuVox in terms of dynamic novel view synthesis and geometry reconstruction.
\input{Fig/4_fig_dynamic_long_seq_tex}
\input{Fig/4_fig_dynamic_free_view_tex}
%
%
\par 
\textbf{Real Scenes.} 
To further evaluate the effectiveness of our method on real scenes with large and non-rigid movement, we select three sequences from the Dynacap~\cite{habermann2021real_DynaCap} dataset. 
Each sequence contains 500 frames with about 50 to 100 camera views for training and about 5 to 10 camera views for testing.
More details are provided in the supp. document.
Tab.~\ref{table:real} summarizes the quantitative comparisons of our method, D-NeRF~\cite{pumarola2020dnerf} and TiNeuVox~\cite{tineuvox}.
Since the real scene datasets do not have ground truth geometry, we can only calculate the PSNR and LPIPS score of novel view synthesis results to evaluate the rendering quality.
For long sequences with 500 frames consisting of challenging movements, D-NeRF and TiNeuVox struggle to reconstruct the dynamic real scenes.
Even when training D-NeRF for 50 hours, our method achieves significantly better scores, taking only 20 seconds per frame.
Also, according to the qualitative evaluation results shown in Fig.~\ref{fig:dynamic_real}, D-NeRF and TiNeuVox show blurred rendering results and inaccurate geometry reconstruction. 
In contrast, our approach produces photo-realistic renderings and detailed geometry.
%
We also conducted experiments on long sequences of 2k and even 20k frames. Here, we present an example with 20K frames in Fig.~\ref{fig:dynamic_long_seq}. More results can be found in the Suppl. materials. 
We provide free view-point results in Fig.~\ref{fig:dynamic_free_viewpoint}, showcasing the sequence in suppl. video at 2:55.
%
%
\subsection{Ablation} \label{sec:ablation}
We first compare our efficient second-order derivative backward computation (Theorem \ref{theorem:mlp_derivative}) implemented in CUDA with a baseline where we automatically compute the second-order derivative in PyTorch~\cite{NEURIPS2019_9015pytorch} using their computational graph. 
As shown in Fig.~\ref{fig:ablation_2order_derivative}, our method achieves a faster speed for second-order derivative backpropagation than the PyTorch implementation in all settings. Moreover, we evaluated our method for calculating MLP's second-order derivatives by eliminating the influence of other factors. The ablation study in Tab.~\ref{table:ablation_pytorch_cmp} further showcases the versatility of NeuS2, demonstrating that it can be easily integrated into other methods for second-order derivatives calculation to greatly bolster training speed and performance.
\par 
Second, we evaluate the performance of the individual components of \textit{NeuS2}: Global Transformation Prediction (GTP) and Progressive Training strategy (PT) on dynamic scenes, as shown in Fig.~\ref{fig:ablation_dynamic} and Tab.~\ref{table:ablation}.
The geometry and appearance reconstruction quality of the full model are better than other ablated models, both, quantitatively and qualitatively.
The ablated model shows noisy holes on the surface and blurred renderings since it gets stuck in local minima during the incremental training, which can be alleviated by the Global Transformation Prediction and Progressive Training.
We also ablated the effectiveness of the Progressive Training Strategy and Eikonal Loss, applied to static scenes within the DTU dataset, as shown in Tab.~\ref{table:ablation_static}.
%
%
We evaluate the impact of predicted transformation accuracy on reconstruction quality, as shown in Suppl. materials.
%
%

%
\input{Fig/5_fig_ablation_2order_derivative_tex}
\input{Fig/5_fig_ablation_dynamic_tex}
\input{Tab/3_tab_ablation_dynamic}
\input{Tab/4_tab_ablation_static}

\input{Tab/5_tab_ablation_pytorch}

%% file: Tab/0_tab_static.tex
%
%

\definecolor{myyellow}{rgb}{1,1, 0.6}
\definecolor{myorange}{rgb}{1, 0.8, 0.6}
\definecolor{myred}{rgb}{1, 0.6, 0.6}

\begin{table}[tb!]
\setlength{\tabcolsep}{3pt}
\begin{center}
\small
\begin{tabular}{lccccc}
\toprule
     &  COLMAP &
 \multicolumn{1}{c}{NeuS}&
 \multicolumn{1}{c}{InstantNGP}&
 \multicolumn{1}{c}{InstantNSR}&
 \multicolumn{1}{c}{\textbf{Ours}}\\
\midrule
CD $\downarrow$ & 1.36 &\colorbox{myyellow}{0.77} & 1.84 & 1.68 & \colorbox{myred}{0.70} \\
PSNR $\uparrow$ &  - & 28.00 & \colorbox{myred}{28.86} & 25.81 & \colorbox{myyellow}{28.82}  \\ 
\midrule
 Runtime & 1 h &
 \multicolumn{1}{c}{8 h} & 
 \multicolumn{1}{c}{5 min} & 
 \multicolumn{1}{c}{8.5 min} &  
 \multicolumn{1}{c}{5 min}
 \\
\bottomrule
\end{tabular}
\end{center}
\vspace{-6pt}
\caption{The quantitative comparison on DTU dataset. 
We color code the \colorbox{myred}{\textbf{best}} and \colorbox{myyellow}{\textbf{second best}} results. 
Our method outperforms other baselines for geometry reconstruction regarding to the Chamfer Distance (CD) and is on par with Instant-NGP of novel view synthesis in terms of PSNR.
}
\label{table:DTU}
\end{table}

%% file: Fig/2_fig_static_tex.tex
\begin{figure*}
	\centering
	\includegraphics[width=1.0\linewidth]{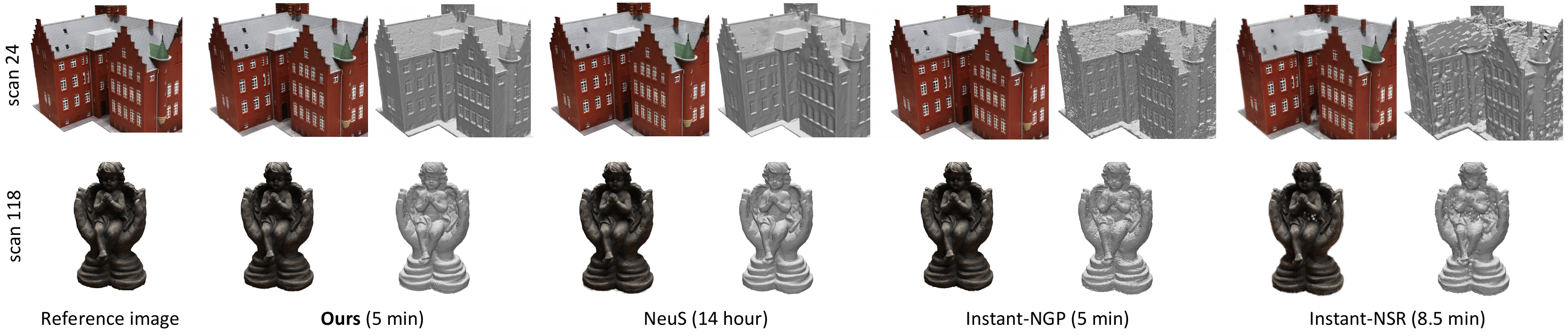}
	\vspace{-15pt}
	\caption
	{
	    Qualitative comparisons on DTU dataset for static scene geometry reconstruction and novel view synthesis.
	    Our method demonstrates high-quality rendering quality, superior to NeuS and comparable to Instant-NGP in terms of complex texture reconstruction.
	    In addition, it outperforms all baselines regarding to 3D geometry reconstruction, with fine details without inducing noise.
	}
	\label{fig:static}
\end{figure*}
%
%

%% file: Tab/1_tab_dynamic_synthetic.tex
\begin{table}[tb!]
\vspace{-9pt}
\small
\begin{center}
\begin{tabular}{lcccccc}
\toprule
     &
 \multicolumn{2}{c}{D-NeRF}&
  \multicolumn{2}{c}{TiNeuVox}&
 \multicolumn{2}{c}{\textbf{Ours}}\\
\midrule
 Dataset & 
 PSNR$\uparrow$&  CD$\downarrow$ & 
  PSNR$\uparrow$&  CD$\downarrow$ & 
 PSNR$\uparrow$&  CD$\downarrow$
 \\
\midrule
Lego & 24.25 &  59.0  & 28.06 & 19.90& \textbf{29.5} & \textbf{17.1}  \\     
Lion & 31.45 &   -  & 31.86 & -  &\textbf{33.60} &  - \\   
Human & 29.33 & 5.73   & 29.00 & 9.21  & \textbf{33.20} & \textbf{1.86}  \\   
\midrule
Runtime &
 \multicolumn{2}{c}{20h} & 
  \multicolumn{2}{c}{1h} & 
 \multicolumn{2}{c}{~1h} \\
\bottomrule
\end{tabular}
\end{center}
\vspace{-6pt}
\caption{Quantitative comparisons on synthetic scenes.
The Chamfer Distance of Lion sequence is omitted since the ground truth geometry is not provided.
Compared to D-NeRF and TiNeuVox, our method achieves much better appearance and geometry reconstruction results.
}
\label{table:synthetics}
\end{table}

%% file: Fig/3_fig_dynamic_synthetic_tex.tex
%
%
\begin{figure}
	\vspace{-10pt}
	\includegraphics[width=\linewidth]{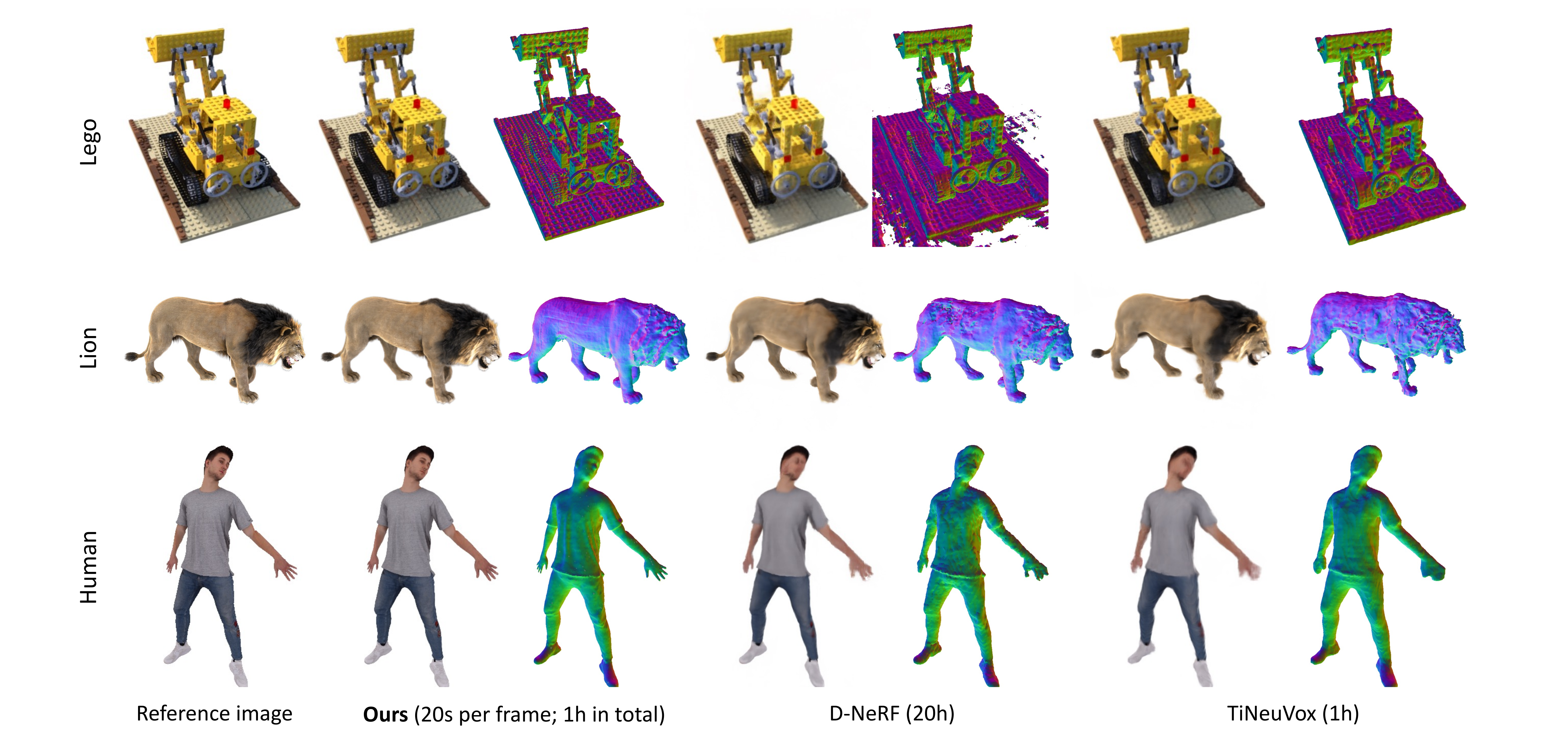}
	\vspace{-14pt}
	\caption
	{
	    Qualitative comparisons of synthetic scene for geometry reconstruction and novel view synthesis.
        Our method produces photo-realistic rendering results and accurate geometry reconstructions with only 20 seconds of training time per frame.
	}
	\label{fig:dynamic_synthetic}
\end{figure}
%
%

%% file: Tab/2_tab_dynamic_real.tex


\begin{table}[tb!]
\vspace{-9pt}
\small
\setlength{\tabcolsep}{3pt}
\begin{center}
\begin{tabular}{lcccccc}
\toprule
     &
 \multicolumn{2}{c}{D-NeRF}& 
 \multicolumn{2}{c}{TiNeuVox}&
 \multicolumn{2}{c}{\textbf{Ours}}\\
\midrule
 Dataset & 
 PSNR$\uparrow$& LPIPS$\downarrow$ & 
 PSNR$\uparrow$& LPIPS$\downarrow$ & 
 PSNR$\uparrow$&  LPIPS$\downarrow$
 \\
\midrule
D1 & 21.48  &  0.122  & 18.17 & 0.159 & \textbf{26.41} &  \textbf{0.036} \\     
D2 & 17.47  &   0.154 & 14.95 & 0.198 & \textbf{27.76} &  \textbf{0.037} \\   
D3 & 20.80 &  0.155  & 14.16 & 0.222 & \textbf{27.25} & \textbf{0.042} \\   
\midrule
Runtime &
 \multicolumn{2}{c}{50h} & 
 \multicolumn{2}{c}{3h} & 
 \multicolumn{2}{c}{~3h}
 \\
 \bottomrule
\end{tabular}
\end{center}
\vspace{-6pt}
\caption{Quantitative comparisons on real scenes.
We found that our method outperforms D-NeRF and TiNeuVox in all metrics.
}
\label{table:real}
\vspace{-10pt}
\end{table}

%% file: Fig/4_fig_dynamic_real_tex.tex
%
%
\begin{figure}
	\vspace{-9pt}
	\includegraphics[width=\linewidth]{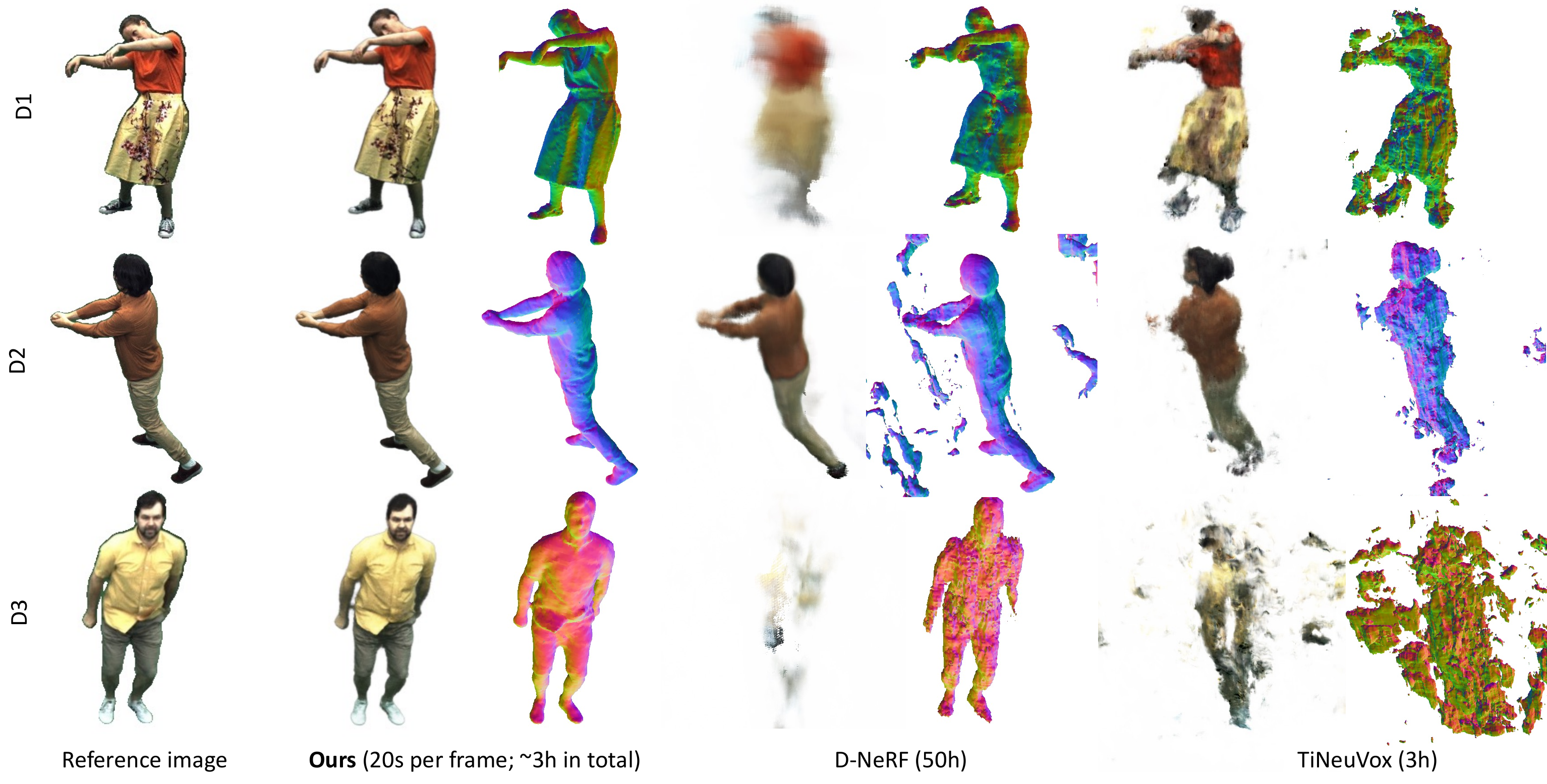}
	\vspace{-14pt}
	\caption
	{
	    Qualitative comparisons of real scenes for geometry reconstruction and novel view synthesis.
	    Our method outperforms D-NeRF and TiNeuVox for both tasks, demonstrating sharp and accurate reconstruction quality, while they fail to handle complex transformations in real scenes.
	}
	\label{fig:dynamic_real}
 	\vspace{-7pt}
\end{figure}
%
%

%% file: Fig/4_fig_dynamic_long_seq_tex.tex
%
%
\begin{figure}[tb!]
	\vspace{-10pt}
    \centering
	\includegraphics[width=1\linewidth]{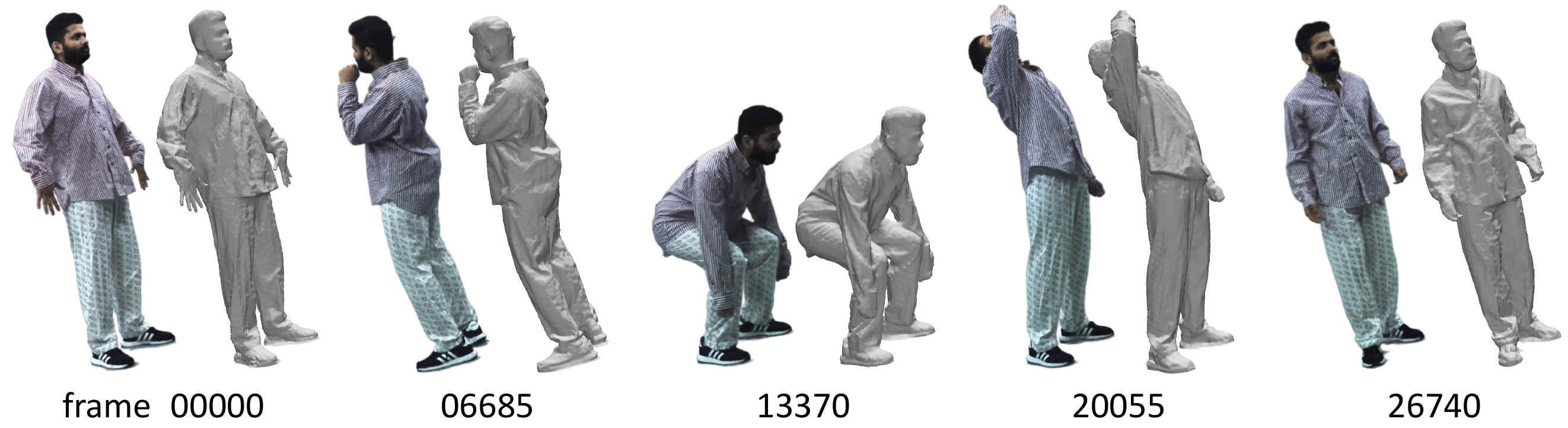}
	\vspace{-10pt}
	\caption
	{
        \footnotesize
	    Long sequence reconstruction results. Our method exhibits robustness even for extremely long   
     sequences, without any performance degradation, e.g., over 20K frames. Video results are provided in the   supplementary material.
	}
    \label{fig:dynamic_long_seq}
    \vspace{-4pt}
\end{figure}
%
%

%% file: Fig/4_fig_dynamic_free_view_tex.tex
%
%
\begin{figure}[tb!]
	\vspace{-5pt}
    \centering
	\includegraphics[width=1\linewidth]{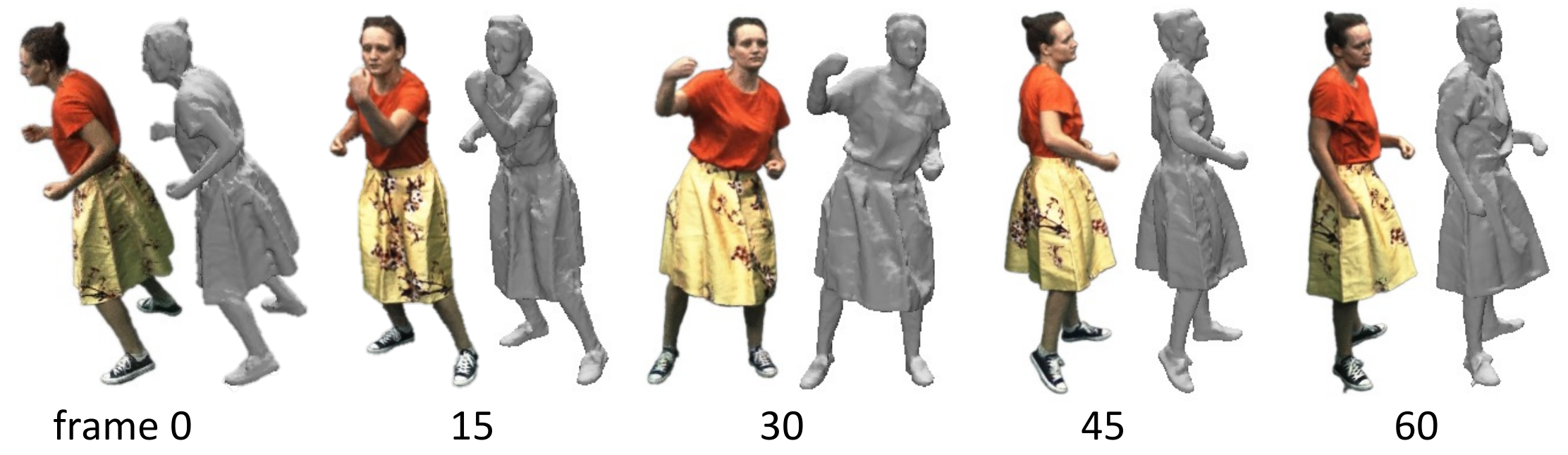}
	\vspace{-10pt}
	\caption
	{
        \footnotesize
	    Free view-point reconstruction results.  Video results are provided in the supplementary material.
	}
    \label{fig:dynamic_free_viewpoint}
    \vspace{-7pt}
\end{figure}
%
%

%% file: Fig/5_fig_ablation_2order_derivative_tex.tex
%
%
\begin{figure}
     \vspace{-12pt}
	\includegraphics[width=\linewidth]{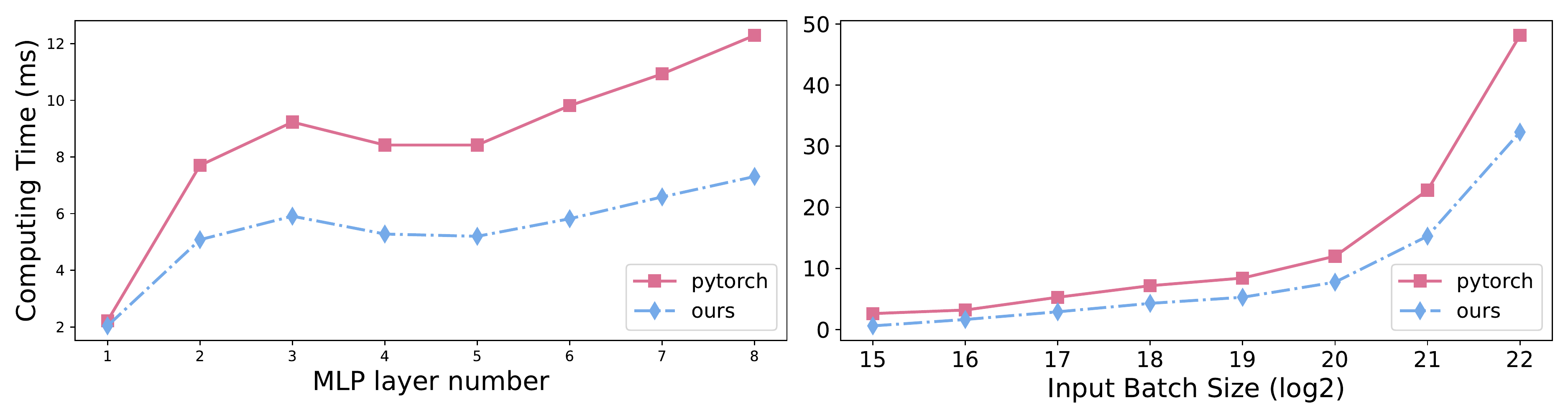}    
    \vspace{-18pt}
	\caption
	{
	    Ablation study for second-order derivative backward computation.
	    We compare the speedup of our 2nd-order derivative with PyTorch across various MLP layer numbers and batchsizes.
	}
	\label{fig:ablation_2order_derivative}
	\vspace{-16pt}
\end{figure}
%
%

%% file: Fig/5_fig_ablation_dynamic_tex.tex
%
%
\begin{figure}
	\centering
  \vspace{1pt}
	\includegraphics[width=0.8\linewidth]{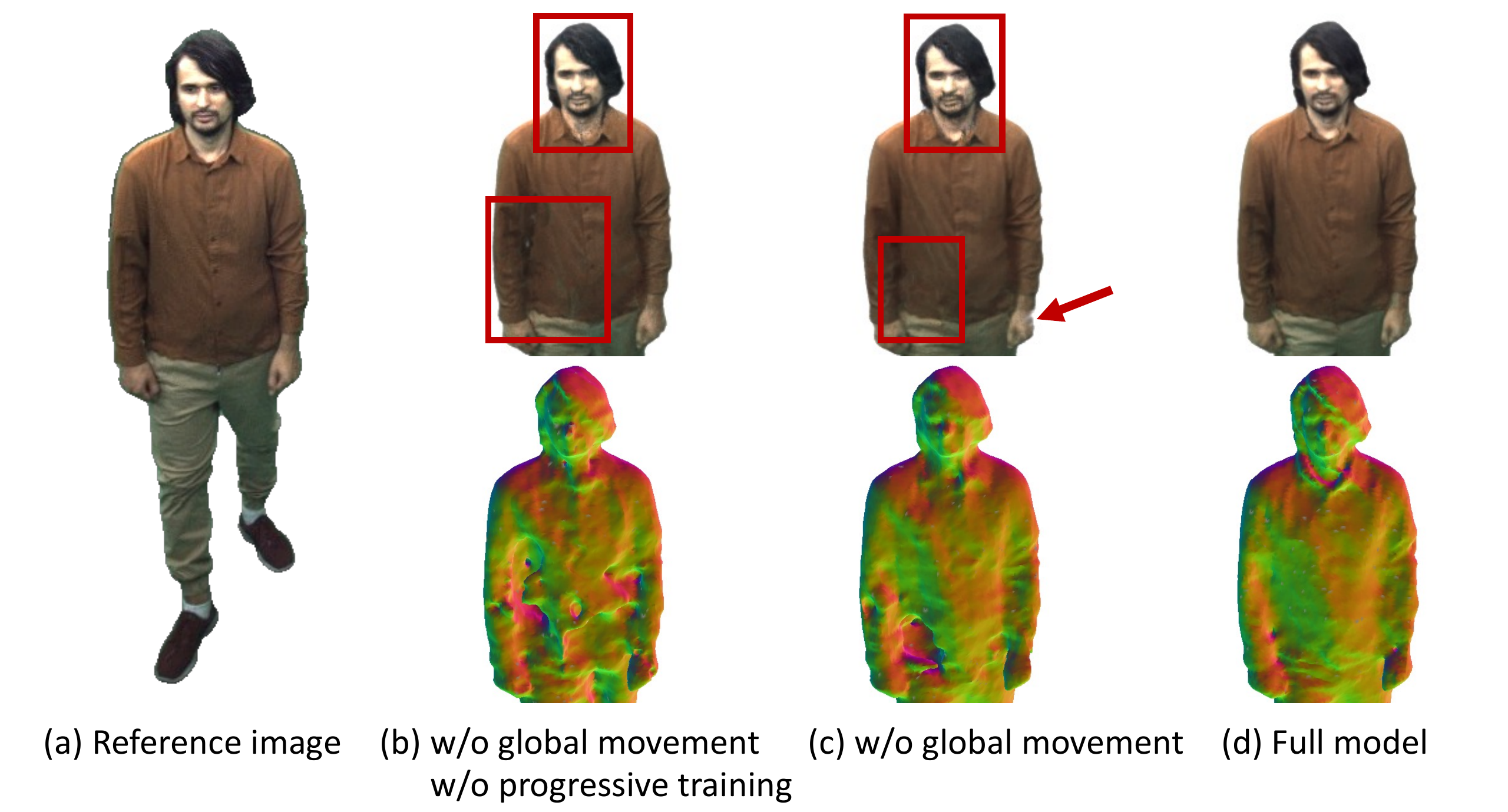}
	\caption
	{
	    Ablation study for Global Transformation Prediction (GTP) and progressive training (PT).
	    Ablated models perform poorly on both novel view synthesis and geometry reconstruction.
	}
	\label{fig:ablation_dynamic}
	\vspace{-7pt}
\end{figure}
%
%

%% file: Tab/3_tab_ablation_dynamic.tex

\begin{table}[tb!]
\vspace{1pt}
\begin{center}
\small
\begin{tabular}{cccc}
\toprule
Methods & w/o GTP + w/o PT & w/o PT & Full model\\
\midrule 
PSNR $\uparrow$ & 27.78 & 27.72 & \textbf{28.18} \\
LPIPS$\downarrow$ & 0.0595 & 0.0559 & \textbf{0.0474}\\
\bottomrule
\end{tabular}
\end{center}
\vspace{-3pt}
\caption{Ablation study of our design choices. Note that, both, the Global Transformation Prediction (GTP) as well as the Progressive Training (PT) improve the overall results.
}
\vspace{-1pt}
\label{table:ablation}
\end{table}

%% file: Tab/4_tab_ablation_static.tex
%
%
\renewcommand{\arraystretch}{0.3}

\definecolor{myyellow}{rgb}{1,1, 0.6}
\definecolor{myorange}{rgb}{1, 0.8, 0.6}
\definecolor{myred}{rgb}{1, 0.6, 0.6}

\begin{table}[tb!]
\vspace{-5pt}
\begin{center}
    \footnotesize
    \begin{tabular}{lccc}
        \toprule
             & 
         \multicolumn{1}{c}{W/O Progressive}&
         \multicolumn{1}{c}{W/O Eikonal}&
         \multicolumn{1}{c}{Full Model}\\
        \midrule
         CD $\downarrow$  & 0.75  & 1.48 & \textbf{0.70} \\
        PSNR $\uparrow$ & 28.75 & 28.73  & \textbf{28.82} \\ 
         \midrule
         Runtime & 
         \multicolumn{1}{c}{5min 40s} & 
         \multicolumn{1}{c}{5min 10s} & 
         \multicolumn{1}{c}{\textbf{5min}}
         \\
        \bottomrule
    \end{tabular}
\end{center}
\vspace{-2pt}

\caption{
\footnotesize Ablation study of Progressive Training and Eikonal loss on DTU. 
}
\label{table:ablation_static}
\vspace{-2pt}
\end{table}

%% file: Tab/5_tab_ablation_pytorch.tex
\definecolor{myyellow}{rgb}{1,1, 0.6}
\definecolor{myorange}{rgb}{1, 0.8, 0.6}
\definecolor{myred}{rgb}{1, 0.6, 0.6}

\newcommand{\tablefirst}[0]{\cellcolor{myred}}
\newcommand{\tablesecond}[0]{\cellcolor{myyellow}}

\setlength{\tabcolsep}{1.8pt}

\begin{table}[tb!]
\vspace{-6pt}
\begin{center}
    \footnotesize
    \begin{tabular}{lccc}
        \toprule
             &  PyTorch's Autograd & NSR's Finite difference & Ours (in PyTorch) \\
        \midrule
        Runtime $\downarrow$ & 10 min & 11.5 min & \textbf{ 7 min}\\
        PSNR $\uparrow$ & 35.3 & 33.4 & \textbf{35.5}\\ 
        \bottomrule
    \end{tabular}
\end{center}
\vspace{-8pt}
\caption{
\footnotesize
Comparison of Pytorch Autograd, Instant-NSR finite differences, and our second-order derivatives. We tested all of them in PyTorch and on the same \href{https://github.com/zhaofuq/Instant-NSR}{codebase}, eliminating other influences, and found that our formulation is superior in terms of speed and accuracy. 
%
%
%
}
\label{table:ablation_pytorch_cmp}
\vspace{-10pt}
\end{table}

%% file: Sec/5_Conclusion.tex
%
%
\vspace{-8pt}
\section{Conclusion} \label{sec:conclusion}
\vspace{-8pt}
\textbf{Limitations.}
While our method reconstructs each frame of a dynamic scene in high quality,
there is no dense surface correspondences across the frames. 
%
A possible way could be to deform a mesh template to fit our learned neural surfaces for each frame, like the mesh tracking used in ~\cite{Collet2015,zhao2022human}. 
Currently, we also need to save the network parameters of each frame (25M). 
As future work, a compression of such parameters encoding the dynamic scene could be explored. 
\par 
We proposed a learning-based method for accurate multi-view reconstruction of both static and dynamic scenes at an unprecedented runtime performance. 
To achieve this, we integrated multi-resolution hash encodings into neural SDF and introduced a simple calculation of the second-order derivatives tailored to our dedicated network architecture. 
To enhance the training convergence, we presented a progressive training strategy to learn multi-resolution hash encodings. 
For dynamic scene reconstruction, we proposed an incremental training strategy with a global transformation prediction component, which leverages the shared geometry and appearance information in two consecutive frames. 

%% file: Sec/6_Acknowledgement.tex
\section*{Acknowledgement}\vspace{-0.17cm}
Christian Theobalt and Marc Habermann were supported by ERC Consolidator Grant 4DReply  (770784). Lingjie Liu was supported by Lise Meitner Postdoctoral Fellowship. 
We would also like to greatly thank Baoquan Chen from Peking University for generously providing computational resources for this research.

%% file: supp/0_intro.tex
%
%
In the following, we provide more details about our method.
First, we present the derivation of Eqs. 5 and 6 (Sec.~\ref{sec:derivation_5_6}), and the proof of Theorem 1 (Sec.~\ref{sec:derivation_theorem}).
%
%
Second, we introduce the dataset (Sec.~\ref{sec:dataset}) we used in the experiments and show more quantitative and qualitative results (Sec.~\ref{sec:add_result}) to further demonstrate the performance of our model.
Finally, we provide implementation details of our method (Sec.~\ref{sec:implement_detail}).

%% file: supp/1_derivation.tex
%
%
\section{Derivation of Equation 5 and Equation 6} \label{sec:derivation_5_6}
%
%
We calculate the second-order derivatives of the hash table parameters $\Omega$ as
%
\begin{equation} \label{supp_eqn:dloss_denc}
\begin{aligned}
\frac{\partial \mathcal{L}}{\partial \Omega}
& = \frac{\partial \mathcal{L}}{\partial \mathbf{n}} \frac{\partial \frac{\partial d}{\partial \mathbf{x}}}{\partial \Omega} \\
&= \frac{\partial \mathcal{L}}{\partial \mathbf{n}} \frac{\partial (\frac{\partial d}{\partial \mathbf{e}} \frac{\partial \mathbf{e}}{\partial \mathbf{x}})}{\partial \Omega} \\
&= \frac{\partial \mathcal{L}}{\partial \mathbf{n}} (
\frac{\partial \mathbf{e}}{\partial \mathbf{x}} \frac{\partial \frac{\partial d}{\partial \mathbf{e}}}{\partial \Omega}
+ \frac{\partial d}{\partial \mathbf{e}} \frac{\partial \frac{\partial \mathbf{e}}{\partial \textbf{x}}}{\partial \Omega}
) \\
&= \frac{\partial \mathcal{L}}{\partial \mathbf{n}} (
\frac{\partial \mathbf{e}}{\partial \mathbf{x}} \frac{\partial \frac{\partial d}{\partial \mathbf{e}}}{\partial \mathbf{e}} \frac{\partial \mathbf{e}}{\partial \Omega}
+ \frac{\partial d}{\partial \mathbf{e}} \frac{\partial \frac{\partial \mathbf{e}}{\partial \mathbf{x}}}{\partial \Omega}
)
\end{aligned}
\end{equation}
%
and the SDF network parameters $\Theta$ as
%
\begin{equation} \label{supp_eqn:dloss_dsdf}
\begin{aligned}
\frac{\partial \mathcal{L}}{\partial \Theta} 
& = \frac{\partial \mathcal{L}}{\partial \mathbf{n}} \frac{\partial \frac{\partial d}{\partial \mathbf{x}}}{\partial \Theta} \\
&= \frac{\partial \mathcal{L}}{\partial \mathbf{n}} \frac{\partial (\frac{\partial d}{\partial \mathbf{e}} \frac{\partial \mathbf{e}}{\partial \mathbf{x}})}{\partial \Theta} \\
&= 
\frac{\partial \mathcal{L}}{\partial \mathbf{n}} (
\frac{\partial \mathbf{e}}{\partial \mathbf{x}} \frac{\partial \frac{\partial d}{\partial \mathbf{e}}}{\partial \Theta} 
+ \frac{\partial d}{\partial \mathbf{e}} \frac{\partial \frac{\partial \mathbf{e}}{\partial \mathbf{x}}}{\partial \Theta}
)
\end{aligned}
\end{equation}

%
%
\section{Derivation of Theorem 1} \label{sec:derivation_theorem}

First we recall the Definition 1.
%
%
%
\begin{definition} \label{supp_def:mlp}

Given a ReLU based MLP $f$ with $L$ hidden layers taking $x \in \mathbb{R}^{d}$ as input, it computes the output $y = H_L g(H_{L-1} \dots g(H_1x))$, where $H_l \in \mathbb{R}^{n_l} \times \mathbb{R}^{n_{l-1}}$, $l\in \{1,\dots,L\}$ is the layer index, and $g$ is the ReLU function. 
We define $P_l^j \in \mathbb{R}^{n_{l-1}} \times \mathbb{R}^1 $ and $S_l^i \in \mathbb{R}^ 1 \times \mathbb{R}^{n_{l}}$ as
%

\begin{equation}
\begin{aligned}
P_l^j &= G_{l} H_{l-1} \dots G_2 H_1^{(\_,j)} \\ 
S_l^i &= H_L^{(i,\_)} G_{L} \dots H_{l+1} G_{l+1}
\end{aligned}
\end{equation}
%
where $H_1^{(\_,j)}$ is the $j$th  column of $H_1$, $H_L^{(i, \_)}$ is the $ i$th row of $H_L$, and 
$G_l=
\begin{cases}
1, H_{l-1}\dots g(H_1x) > 0 \\
0, \text{ otherwise} 
\end{cases}$.
\end{definition}
Now the second-order derivatives of a ReLU-based MLP with respect to its input and intermediate layers can be defined as follows.
%
\begin{theorem} [Second-order derivative of ReLU-based MLP] \label{supp_theorem:mlp_derivative}
Given a ReLU based MLP $f$ with $L$ hidden layers with the same definition in Definition~\ref{def:mlp}. The second-order derivative of the MLP $f$ is:
\begin{equation} \label{supp_eqn:mlp_2ord_derivative_formula}
\frac{\partial \frac{\partial y}{\partial x}_{(i,j)}}{\partial H_l} = (P_l^j S_l^i)^T
,\qquad\frac{\partial^2 y}{\partial \mathbf{x}^2} = 0
\end{equation}
where $\frac{\partial y}{\partial x}_{(i,j)}$ is the matrix element (i,j) of $\frac{\partial y}{\partial x}$, and $S_l^i$ and $P_l^j$ are defined in Definition \ref{def:mlp}.
\end{theorem}
%
\textbf{Proof} \label{prf:ml_derivative}
First, by applying the chain rule, we have
    \begin{equation}
    \begin{aligned}
        \frac{\partial y}{\partial x}
        &= \frac{\partial H_L g(H_{L-1}P_{L-1})}{\partial x} \\
        &= H_L \frac{\partial g(H_{L-1}P_{L-1})}{\partial x} \\
        &= H_L \frac{\partial g}{\partial H_{L-1}P_{L-1}} H_{L-1} \frac{\partial P_{L-1}}{\partial x} \\
        &= H_L G_L H_{L-1} \frac{\partial P_{L-1}}{\partial x} \\
        &= H_L G_L H_{L-1} G_{L-1} H_{L-2} \frac{\partial P_{L-2}}{\partial x} \\
        &= H_L G_L H_{L-1} \cdots G_2 H_1
    \end{aligned}
    \end{equation}
    And, the element $(i, j)$ of $\frac{\partial y}{\partial x} \in \mathbb{R}^{n_{L}} \times \mathbb{R}^{n_{1}} $ is
    \begin{equation}
        \frac{\partial y}{\partial x}_{(i,j)} = H_L^{(i, \_)} G_{L} \cdots G_2 H_1^{(\_, j)}.
    \end{equation}
    We can then calculate
    \begin{equation}
        \begin{aligned}
            \frac{\partial \frac{\partial y}{\partial x}_{(i,j)}}{\partial H_l} 
            = (\frac{\partial \frac{\partial y}{\partial x}_{(i,j)}}{\partial H_l P_l^j})^\top \frac{\partial H_l P_l^j}{\partial H_l}.
        \end{aligned}
    \end{equation}
    On the one hand, since $P_l^j$ is independent of $H_l$, we have
    \begin{equation}
        \begin{aligned}
            \frac{\partial H_l P_l^j}{\partial H_l} = (P_l^j)^\top.
        \end{aligned}
    \end{equation}
    On the other hand, we have
    \begin{equation}
        \begin{aligned}
            \frac{\partial \frac{\partial y}{\partial x}_{(i,j)}}{\partial H_l P_l^j} 
            = \frac{\partial \frac{\partial y}{\partial x}_{(i,j)}}{\partial H_{L-1} P_{L-1}^j} \frac{\partial H_{L-1} P_{L-1}^j}{\partial H_{L-2} P_{L-2}^j} \cdots \frac{\partial H_{l+1} P_{l+1}^j}{\partial H_{l} P_{l}^j}.
        \end{aligned}
    \end{equation}
    we can further derive that  
    \begin{equation}
        \frac{\partial \frac{\partial y}{\partial x}_{(i,j)}}{\partial H_{L-1} P_{L-1}^j} = H_L^{(i,\_)} G_{L},
    \end{equation}
    and
    \begin{equation}
    \begin{aligned}
        \frac{\partial H_{l} P_{l}^j}{\partial H_{l-1} P_{l-1}^j} 
        &= \frac{\partial H_{l} G_{l} H_{l-1} P_{l-1}^j}{\partial H_{l-1} P_{l-1}^j} \\
        &= H_{l} G_{l} + H_{l} \frac{\partial G_{l}}{\partial H_{l-1} P_{l-1}^j} \\
        &= H_{l} G_{l},
    \end{aligned}
    \end{equation}
    since $\frac{\partial G_{l}}{\partial H_{l-1} P_{l-1}^j} = 0$. Thus we have
    \begin{equation}
    \begin{aligned}
        \frac{\partial \frac{\partial y}{\partial x}_{(i,j)}}{\partial H_l P_l^j} 
        = H_L^{(i,\_)} G_{L} H_{L-1} \cdots H_{l+1} G_{l+1} = S_l^{i}.
    \end{aligned}
    \end{equation}
    Consequently,
    \begin{equation}
        \begin{aligned}
            \frac{\partial \frac{\partial y}{\partial x}_{(i,j)}}{\partial H_l} 
            = (\frac{\partial \frac{\partial y}{\partial x}_{(i,j)}}{\partial H_l P_l^j})^\top \frac{\partial H_l P_l^j}{\partial H_l} 
            = (S_l^{i})^\top (P_l^j)^\top.
        \end{aligned}
    \end{equation}
    Second, we define $x_l \in \mathbb{R}^{n_{l}}$, $l\in \{1,\dots,L-1\}$ as $x_l = g(H_{l-1}x_{l-1})$, specifically $x_1 = g(H_1 x)$.
    Then we have 
    \begin{equation}
        y = H_L x_{L-1}.
    \end{equation}
    Then we calculate
    \begin{equation}
        \begin{aligned}
            \frac{\partial^2 y}{\partial \mathbf{x}^2}
            &= \frac{\partial \frac{\partial y}{\partial x_1} \frac{\partial x_1}{\partial x}}{\partial x}
            &= \frac{\partial \frac{\partial y}{\partial x_1}}{\partial x} \frac{\partial x_1}{\partial x} + \frac{\partial \frac{\partial x_1}{\partial x}}{\partial x} \frac{\partial y}{\partial x_1}.
        \end{aligned}
    \end{equation}
    Note that $\frac{\partial \frac{\partial x_1}{\partial x}}{\partial x} = 0$.
    So we have
    \begin{equation}
        \begin{aligned}
            \frac{\partial^2 y}{\partial \mathbf{x}^2}
            &= \frac{\partial \frac{\partial y}{\partial x_1}}{\partial x} \frac{\partial x_1}{\partial x} \\
            &= \frac{\partial^2 y}{\partial x_{1}^2} (\frac{\partial x_1}{\partial x})^2 \\
            &= \frac{\partial^2 y}{\partial x_{L-1}^2} (\frac{\partial x_{L-1}}{\partial x_{L-2}})^2 \cdots (\frac{\partial x_1}{\partial x})^2.
        \end{aligned}
    \end{equation}
    Since $y= H_L x_{L-1}$, we have $\frac{\partial^2 y}{\partial x_{L-1}^2} = 0$.
    Thus we get
    \begin{equation}
        \begin{aligned}
            \qquad\frac{\partial^2 y}{\partial \mathbf{x}^2}
            &= 0
        \end{aligned}
    \end{equation}

%% file: supp/1.5_concurrent_works.tex
%
%

%% file: supp/2_dataset.tex
%
%
\section{Dataset}\label{sec:dataset}
%
%

%
%
\textbf{Dataset for Static Scene Reconstruction.}
For static scene reconstruction, we use 15 scenes from the DTU dataset~\cite{DTUdataset}, same as those used in NeuS~\cite{wang2021neus}, with a wide variety of materials, appearance and geometry.
These scenes involve challenging cases for reconstruction algorithms, such as non-Lambertian surfaces and fine structures.
Each scene contains 49 or 64 images with an image resolution of $1600 \times 1200$.
We split each scene into training and testing parts following NeuS~\cite{wang2021neus}.
Specifically, we set images indexed at 8, 13, 16, 21, 26, 31, 34 and 56 if available for testing and the other images for training.
We train and test each scene with foreground masks provided by IDR~\cite{idr_yariv2020multiview}.

%
%
\textbf{Dataset for Dynamic Synthetic Scene Reconstruction.}
We use three synthetic scenes with various types of deformations and motions to evaluate our method. The Lego scene is shared by NeRF~\cite{mildenhall2020nerf} in form of Blender~\cite{Blender} file. We transform the Lego object into different poses and positions and render images at the resolution of 400 $\times$ 400 in Blender. The Lego scene contains 150 frames with 40 training camera views and 40 test camera views. The human scene is provided by RenderPeople\cite{Renderpeople}. We render images at the resolution of $512 \times 512$ following~\cite{saito2019pifu}, and the whole sequence contains 100 frames with 48 camera views for training and 12 camera views for testing.
The Lion scene is shared by Artemis ~\cite{luo2022artemis}, which has 177 frames with 30 camera views for training and 6 camera views for testing.
%
%
\textbf{Dataset for Dynamic Real Scene Reconstruction.}
We select three sequences from the Dynacap dataset~\cite{habermann2021real_DynaCap}, denoted as D1, D2 and D3, for real-scene reconstruction.
These sequences are captured under a dense camera setup at a resolution of $1285 \times 940$. 
We crop the images with a 2D bounding box which is estimated from the foreground masks to obtain the target images at a resolution of $512 \times 512$.
For each sequence, we choose 500 frames containing large movements for evaluation to show our advantages (D1: 17,760 to 18,260, D2: 6,095 to 6,595, D3: 3,450 to 3,950).
The D1 sequence has 94 camera views, from which we pick 9 camera views (7, 17, 27, 37, 47, 57, 67, 77, 87) for testing and the rest of the views for training.
The D2 sequence has 50 camera views, from which we pick 5 camera views (7, 17, 27, 37, 47) for testing and the rest of the views for training.
The D3 sequence has 101 camera views, from which we pick 10 camera views (7, 17, 27, 37, 47, 57, 67, 77, 87, 97) for testing and the rest of the views for training.
We train and test each scene with foreground masks provided by Dynacap~\cite{habermann2021real_DynaCap}.
We choose a sequence of a human walking around in the D1 sequence, which contains 200 frames, to conduct the ablation study.

%% file: supp/3_additional_results.tex
%
%
\section{Additional Result} \label{sec:add_result}
\input{supp_tab/0_tab_static}
\input{supp_fig/1_fig_static_supp_tex}
\input{supp_tab/1_tab_voxurf}
\input{supp_tab/2_tab_dynamic_ngp}
%
%
\textbf{Video Results.}
We provide a supplementary video to better demonstrate the qualitative results of our method.
We highly encourage the readers to check our video.

%
%
\textbf{Static Scene Reconstruction.}\label{sec:results_static}
In Tab.~\ref{table:DTU_supp}, we provide the per-scene breakdown analysis of the quantitative comparisons on the DTU dataset presented in the main paper (Tab. 1).
We also present additional qualitative comparisons on the DTU dataset in Fig.~\ref{fig:static_supp}.
Additionally, we provide a quantitative comparison with a concurrent work, Voxurf~\cite{Voxurf}, in Tab.~\ref{table:voxurf_supp} under its experimental settings.
\par
%
%
\textbf{Dynamic Scene Reconstruction.}\label{sec:results_dynamic}
In Tab.~\ref{table:supp_synthetics}, we present the quantitative comparison between our method and Instant-NGP~\cite{muller2022instant_ngp} for synthetic scenes.

\par
%
%
\textbf{Effect of the accuracy of Global Transformation Prediction.}
For our global Transformation Prediction compoment, a few minor inaccuracies will not have a significant impact on the final result, since our incremental training strategy can further fine-tune the model parameters to compensate for these errors.
To illustrate this statement, we conduct an experiment on one sequence in the DynaCap. 
We use the $R$ and $T$ of the SMPL obtained by EasyMocap as ground truth, and then add different levels of noises to get different accuracy levels of $R$ and $T$. 
The results in Fig. ~\ref{fig:rebuttal_transform_acc} demonstrate that our model is rather robust for the predicted rotation $R$ and transition $T$.  
The reconstruction performance only drops when the predicted transition and rotation are very inaccurate.
\input{supp_fig/2_fix_transform_acc_tex}
%
%

%
%

%

%% file: supp_tab/0_tab_static.tex
\definecolor{myyellow}{rgb}{1,1, 0.6}
\definecolor{myorange}{rgb}{1, 0.8, 0.6}
\definecolor{myred}{rgb}{1, 0.6, 0.6}

\setlength{\tabcolsep}{3pt}
\begin{table*}[tb!]
\renewcommand\arraystretch{1.30}
\begin{center}
\caption{Quantitative comparisons on the DTU dataset. 
We color code the \colorbox{myred}{\textbf{best}} and \colorbox{myyellow}{\textbf{second best}} results. 
Our method outperforms other baselines for geometry reconstruction regarding the Chamfer Distance (CD) and is on par with Instant-NGP of novel view synthesis in terms of PSNR. 
}
\vspace{5pt}
\label{table:DTU_supp}
\begin{tabular}{c|c|cc|cc|cc|cc}
\hline
     &  COLMAP &
 \multicolumn{2}{|c}{NeuS}&
 \multicolumn{2}{|c}{Instant-NGP}&
 \multicolumn{2}{|c}{Instant-NSR}&
 \multicolumn{2}{|c}{Ours}\\
\hline
 Runtime & 1h &
 \multicolumn{2}{|c}{8 h} & 
 \multicolumn{2}{|c}{5 min} & 
 \multicolumn{2}{|c}{8.5 min} & 
 \multicolumn{2}{|c}{5 min}
 \\
\hline
 ScanID & CD $\downarrow$ &
 PSNR$\uparrow$&  CD $\downarrow$ & 
 PSNR$\uparrow$&  CD$\downarrow$ & 
 PSNR$\uparrow$&  CD$\downarrow$ & 
 PSNR$\uparrow$&  CD$\downarrow$
 \\
\hline
scan24 & \colorbox{myyellow}{0.81} &26.49 & 0.83   & \colorbox{myyellow}{28.32} & 1.68 & 23.86 & 2.86   & \colorbox{myred}{28.44} & \colorbox{myred}{0.56}  \\     
scan37 & 2.05 &26.17 & \colorbox{myyellow}{0.98}   & \colorbox{myred}{27.19} & 1.93 & 24.97 & 2.81 & \colorbox{myyellow}{27.14} & \colorbox{myred}{0.76}  \\   
scan40 & 0.73 &27.66 & \colorbox{myyellow}{0.56}   & \colorbox{myred}{30.45} & 1.57 & 25.3	& 2.09   & \colorbox{myyellow}{29.70} & \colorbox{myred}{0.49}  \\   
scan55 & 1.22 &27.78 & \colorbox{myred}{0.37}   & \colorbox{myred}{29.81} & 1.16 & 25.43 & 0.81   & \colorbox{myyellow}{29.67} & \colorbox{myred}{0.37}  \\   
scan63 & 1.79 &30.63 & \colorbox{myyellow}{1.13}   & \colorbox{myyellow}{31.22} & 2.00 & 29.52 & 1.65   & \colorbox{myred}{31.75} & \colorbox{myred}{0.92}  \\   
scan65 & 1.58 &27.42 & \colorbox{myred}{0.59}   & \colorbox{myyellow}{27.78} & 1.56 & 26.17 & 1.39   & \colorbox{myred}{27.83} & \colorbox{myyellow}{0.71}  \\   
scan69 & 1.02 &\colorbox{myred}{25.83} & \colorbox{myred}{0.60} & 24.79 & 1.81 & 22.93 & 1.47    & \colorbox{myyellow}{24.84} & \colorbox{myyellow}{0.76}  \\   
scan83 & 3.05 &30.00  & \colorbox{myyellow}{1.45}   & \colorbox{myyellow}{31.23} & 2.33  & 26.72 & 1.67  & \colorbox{myred}{31.24} & \colorbox{myred}{1.22}  \\   
scan97 & 1.40 &26.40 &  \colorbox{myred}{0.95}  & \colorbox{myred}{26.96} & 2.16 & 25.94 & 2.47   & \colorbox{myyellow}{26.86} & \colorbox{myyellow}{1.08}  \\   
scan105 & 2.05 &29.63 & \colorbox{myyellow}{0.78}   & \colorbox{myred}{30.62} & 1.88 & 27.71 & 1.12    & \colorbox{myyellow}{30.57} & \colorbox{myred}{0.63}  \\   
scan106 & 1.00 &\colorbox{myyellow}{25.87} &  \colorbox{myred}{0.52}   & 25.62 & 1.76 & 23.12 & 1.22   & \colorbox{myred}{26.05} & \colorbox{myyellow}{0.59}  \\   
scan110 & \colorbox{myyellow}{1.32} &\colorbox{myyellow}{28.82} & 1.43   & 28.6 & 2.32 & 25.44 & 2.30   & \colorbox{myred}{28.93} & \colorbox{myred}{0.89}  \\   
scan114 & 0.49 &28.80 & \colorbox{myred}{0.36}   & \colorbox{myred}{29.5} & 1.86 & 26.7	& 0.98   & \colorbox{myyellow}{28.98} & \colorbox{myyellow}{0.40}  \\   
scan118 & 0.78 &27.36 & \colorbox{myred}{0.45}   & \colorbox{myred}{27.91} & 1.80 & 25.13 & 1.41   & \colorbox{myyellow}{27.82} & \colorbox{myyellow}{0.48}  \\   
scan122 & 1.17 &31.19 & \colorbox{myred}{0.45}   & \colorbox{myred}{32.93} & 1.72 & 28.19 & 0.95   & \colorbox{myyellow}{32.48} & \colorbox{myyellow}{0.55}  \\   
\hline
mean &  1.36 & 28.00 & \colorbox{myyellow}{0.77}  & \colorbox{myred}{28.86} & 1.84 & 25.81 & 1.68   & \colorbox{myyellow}{28.82} & \colorbox{myred}{0.70}  \\

\hline
\end{tabular}
\end{center}
\end{table*}

%% file: supp_fig/1_fig_static_supp_tex.tex
%
%
\begin{figure*}
	\centering
	\includegraphics[width=\linewidth]{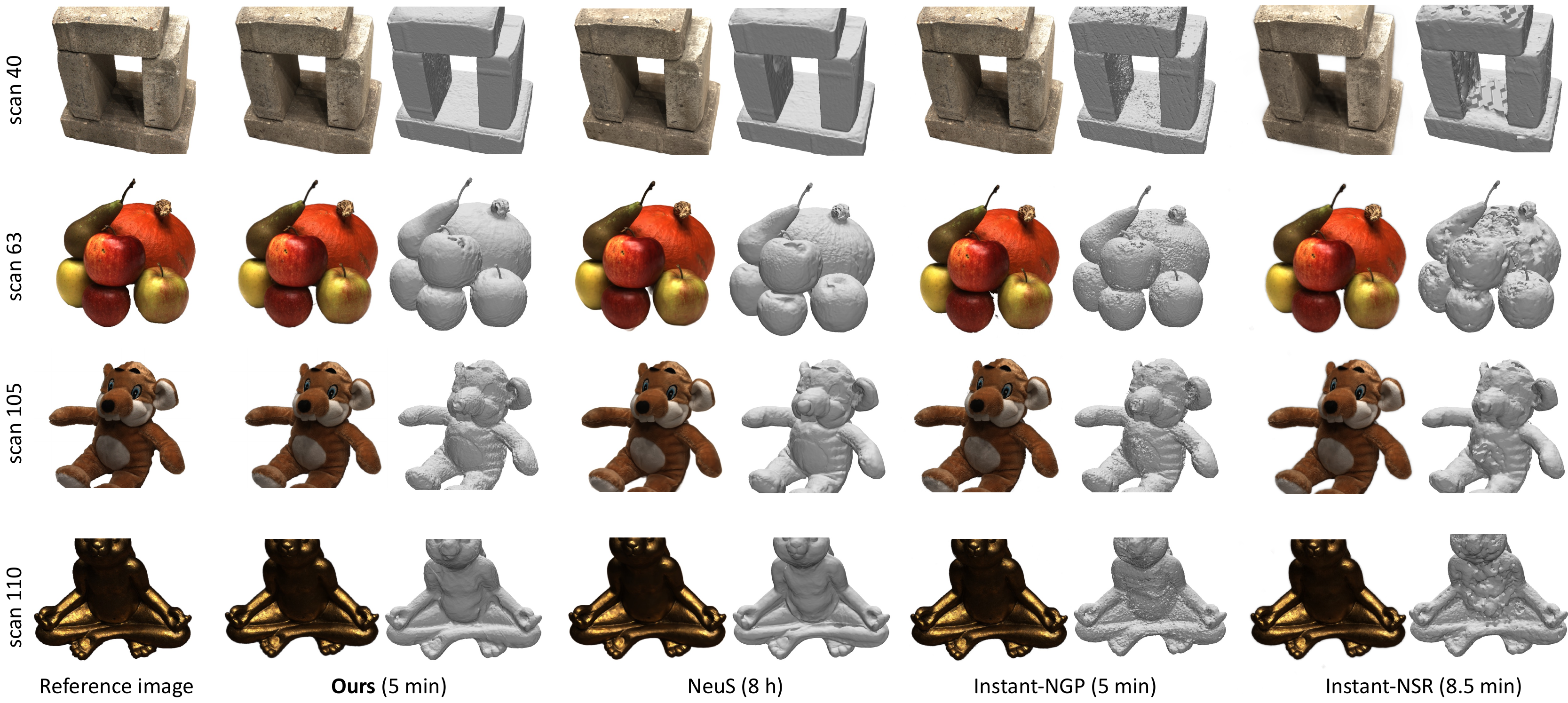}
	\caption
	{
	    Qualitative comparisons on the DTU dataset for static scene geometry reconstruction and novel view synthesis.
	    Our method demonstrates high-quality rendering quality, superior to NeuS and  Instant-NSR, and comparable to Instant-NGP in terms of complex texture reconstruction.
	    In addition, it outperforms all the baselines regarding 3D geometry reconstruction, with fine details without inducing noise.
	}
	\label{fig:static_supp}
\end{figure*}
%
%

%% file: supp_tab/1_tab_voxurf.tex
\definecolor{myyellow}{rgb}{1,1, 0.6}
\definecolor{myorange}{rgb}{1, 0.8, 0.6}
\definecolor{myred}{rgb}{1, 0.6, 0.6}

\setlength{\tabcolsep}{4pt}
\begin{table}[tb!]
\renewcommand\arraystretch{1.25}
\begin{center}
\caption{
Quantitative comparison between our method and Voxurf~\cite{Voxurf}, using the DTU dataset and the experimental settings employed by Voxurf.} 
\vspace{5pt}
\label{table:voxurf_supp}
\begin{tabular}{c|cc|cc}
\hline
     &  
 \multicolumn{2}{|c}{Voxurf}&
 \multicolumn{2}{|c}{Ours}\\
\hline
 Runtime & 
 \multicolumn{2}{|c}{16 min} & 
 \multicolumn{2}{|c}{\textbf{5 min}}
 \\
\hline
 ScanID & 
 PSNR$\uparrow$&  CD$\downarrow$ & 
 PSNR$\uparrow$&  CD$\downarrow$
 \\
\hline
scan24 & 27.89 &  0.65  & \textbf{28.44} & \textbf{0.56}  \\     
scan37 & \textbf{26.90} &  \textbf{0.74}  & 26.53 & 0.76  \\   
scan40 & 28.81 &  \textbf{0.39}  & \textbf{29.70} & 0.49  \\   
scan55 & 31.02 &  \textbf{0.35}  & \textbf{31.47} & 0.37  \\   
scan63 & \textbf{34.38} &  0.96  & 33.74 & \textbf{0.92}  \\   
scan65 & \textbf{31.48} &  \textbf{0.64} & 30.99 & 0.71  \\   
scan69 & \textbf{30.13} &  0.85  & 28.77 & \textbf{0.76}  \\   
scan83 & \textbf{37.43} &  1.58  & 36.78 & \textbf{1.22}  \\   
scan97 & \textbf{28.35} &  \textbf{1.01} & 28.24 & 1.08  \\   
scan105 & 32.94 &  0.68 & \textbf{33.30} & \textbf{0.63}  \\   
scan106 & \textbf{34.17} &  0.60 & {33.91} & \textbf{0.59}  \\   
scan110 & 32.70 &  1.11 & \textbf{34.50} & \textbf{0.89}  \\   
scan114 & 30.97 &  \textbf{0.37} & \textbf{31.14} & {0.40}  \\   
scan118 & \textbf{37.24} &  \textbf{0.45} & {37.17} & {0.48}  \\   
scan122 & \textbf{37.97} & \textbf{0.47} & {37.41} & {0.55}  \\   
\hline
mean &  \textbf{32.16} &  0.72 & {32.14} & \textbf{0.70}  \\

\hline
\end{tabular}
\end{center}
\end{table}

%% file: supp_tab/2_tab_dynamic_ngp.tex
\begin{table}[tb!]
\begin{center}
\begin{tabular}{lcccc}
\toprule
     &
 \multicolumn{2}{c}{Instant-NGP}&
 \multicolumn{2}{c}{\textbf{Ours}}\\
\midrule
 Dataset & 
 PSNR$\uparrow$&  CD$\downarrow$ & 
 PSNR$\uparrow$&  CD$\downarrow$
 \\
\midrule
Lego & 29.05 &  37.19  & \textbf{29.5} & \textbf{17.1}  \\     
Lion & 33.09 &   -  &\textbf{33.60} &  - \\   
Human & \textbf{36.18} & 4.48  & 33.20 & \textbf{1.86}  \\   
\midrule
Runtime &
  \multicolumn{2}{c}{~1h} & 
 \multicolumn{2}{c}{~1h} \\
\bottomrule
\end{tabular}
\end{center}
\caption{
Quantitative comparison between our method and Instant-NGP~\cite{muller2022instant_ngp} on synthetic scenes.
The Chamfer Distance of Lion sequence is omitted since the ground truth geometry is not provided.
}
\label{table:supp_synthetics}
\end{table}

%% file: supp_fig/2_fix_transform_acc_tex.tex
%
%
\begin{figure}[tb!]
    \centering
	\includegraphics[width=1\linewidth]{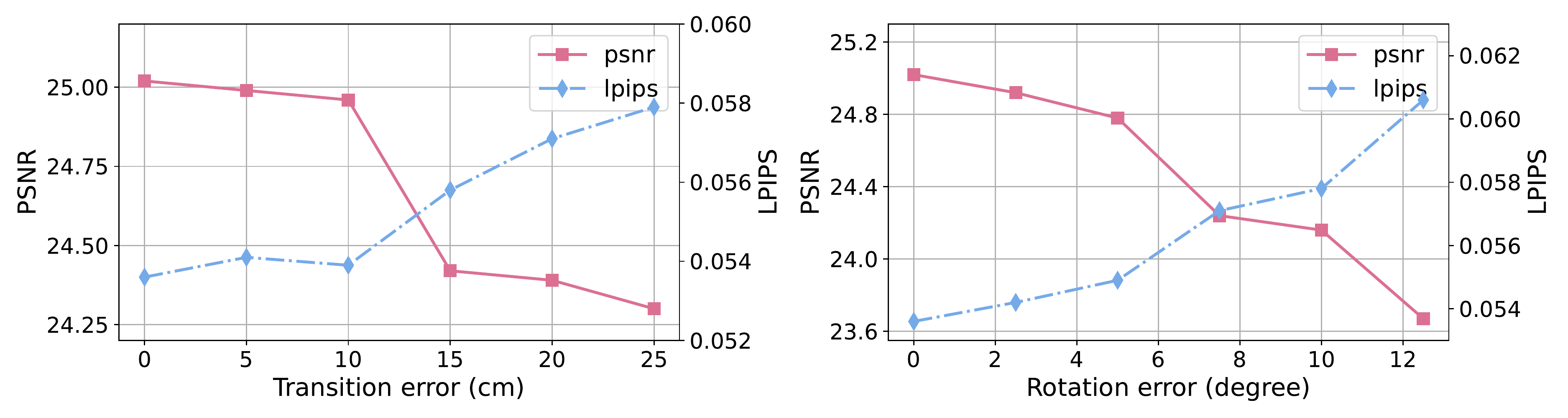}
	%
	\caption
	{
	    Influence of the predicted transformation accuracy to the reconstruction quality. We use the rotation $R$ and transition $T$ of  SMPL as ground truth, and then add different levels of noise for different levels of accuracy.
	}
	\label{fig:rebuttal_transform_acc}
	%
        %
\end{figure}
%
%

%% file: supp/4_implemental_details.tex
%
%
\section{Implementation Details} \label{sec:implement_detail}
%
\input{supp_fig/0_fig_network_arch}
%
%
\subsection{Baselines}
\textbf{COLMAP~\cite{schoenberger2016sfm,schoenberger2016mvs}.}
We directly refer to COLMAP's results on the DTU dataset reported in NeuS\cite{wang2021neus}.
\par
%
%
\textbf{NeuS~\cite{wang2021neus}.}
The Chamfer Distance scores of NeuS shown in the paper are directly referred  to the results reported in the original paper.
The geometry reconstruction results are produced using the officially released pre-trained models\footnote{https://github.com/Totoro97/NeuS\label{neus}} with mask supervision.
The PSNR scores and novel view synthesis results are obtained by training the officially released code\textsuperscript{\ref{neus}}
 on the DTU training dataset with mask supervision, and testing it on the DTU testing dataset.
\par
%
%
\textbf{Instant-NGP~\cite{muller2022instant_ngp}.}
We use the officially released code \footnote{https://github.com/NVlabs/instant-ngp} to train the model on the DTU dataset for 50k iterations. The training takes about 5 minutes.
For dynamic scenes, we train the model separately for each frame from scratch, and limit the training time to 20 seconds which is consistent with our method.
\par
%
%
\textbf{Instant-NSR~\cite{zhao2022human}.}
The results on DTU dataset are provided by the authors.
\par
%
%
\textbf{Voxurf~\cite{Voxurf}.}
The results on DTU dataset are refered to the original paper.
\par
%
%
\textbf{D-NeRF~\cite{pumarola2020dnerf}.}
We use the officially released code \footnote{https://github.com/albertpumarola/D-NeRF} to train the model for 800k iterations.
The training time of D-NeRF on real scenes is longer than on synthetic scenes, 50 and 20 hours, respectively.
This is because the length of real sequences is longer than that of the synthetic sequences, which are around 500 frames and 150 frames, respectively.
Moreover the number of camera views for real scenes is greater than for synthetic scenes.
For long sequences with dense camera views, the model cannot upload all the images at once due to the GPU memory limitation, so extra time is needed to load the images during training.
\par
%
%
\textbf{TiNeuVox~\cite{tineuvox}.}
We use the officially released code \footnote{https://github.com/hustvl/TiNeuVox} and train the model for 80k and 150k iterations on synthetic scenes and real scenes, respectively. Due to the same reason mentioned above, extra time is needed during the training on real scenes. The training time of TiNeuVox on real scenes is longer than on synthetic scenes, 3 and 1 hours, respectively.
\par
%
%
\subsection{Network Architecture}
As shown in Fig \ref{fig:network_arch}, the network architecture of \textit{NeuS2} consists of the following components:
(a) a multi-resolution hash grid with 14 levels of different resolutions ranging from 16 to 2048;
(b) an SDF network modeled by a 1-layer MLP with 64 hidden units;
(c) an RGB network modeled by a 2-layer MLP with 64 hidden units.

%
%
\subsection{Training Details}
\textbf{Unbiased Volume Rendering.}
To render an image, we apply the unbiased volume rendering of NeuS~\cite{wang2021neus}.
That is, we first transform the signed distance field into a volume density field $\phi_s(f(\textbf{x}))$, where $\phi_s(x)=se^{-sx}/(1+e^{-sx})^2$ is the \textit{logistic density distribution}, which is the derivative of the Sigmoid function $\Phi_s(x)=1/(1+e^{-sx})$ and $s$ is a learnable parameter.
Next, we construct an unbiased weight function in the volume rendering equation. 
Specifically, for each pixel of an image, we sample $n$ points $\{\mathbf{\mathbf{p}(t_i)}=\mathbf{o}+t_i\mathbf{v}|i=0,1,\dots, n-1\}$ along its camera ray, where $\mathbf{o}$ is the center of the camera and $\mathbf{v}$ is the view direction.
By accumulating the SDF-based densities and colors of the sample points, we can compute the color $\hat{C}$ of the ray with the same approximation scheme as used in NeRF~\cite{mildenhall2020nerf} as
%
\begin{equation}
\hat{C}(\mathbf{o}, \mathbf{v}) = \sum_{i=0}^{n-1} T({t_i})\alpha({t_i})c(\mathbf{p}({t_i}), \mathbf{v})
\label{eqn:neus_volume_rendering}
\end{equation}
%
where $T(t_i)$ is the discrete accumulated transmittance defined by $T(t_i)=\prod_{j=0}^{i-1}(1-\alpha(t_j))$, and $\alpha(t_i)$ is a discrete density value defined by
%
\begin{equation} \label{eqn:neus_rendering_alpha}
\alpha(t_i) = \max\Big(\frac{\Phi_s(f(p(t_{i})))-\Phi_s(f(p(t_{i+1})))}{\Phi_s(f(p(t_{i})))}, 0\Big).
\end{equation}
%
As the rendering process is differentiable, we can learn the signed distance field $f$ and the radiance field $c$ from the multi-view images. 
\par
%
%
\textbf{Ray Marching Strategy.}
We adopt a ray marching acceleration strategy used in Instant-NGP\cite{muller2022instant_ngp}.
That is, we maintain an occupancy grid that roughly marks each voxel grid as empty or non-empty.
The occupancy grid can effectively guide the marching process by preventing sampling in empty spaces and, thus, accelerate the volume rendering process.
We periodically update the occupancy grid based on the SDF value predicted by our model. In detail, we first use the SDF value $d$ to calculate the density value $\phi_s(d)$ for each grid, where $\phi_s(x)=se^{-sx}/(1+e^{-sx})^2$ is the \textit{logistic density distribution} and $s$ is a learnable parameter. We then use the density value to update the occupancy grid in the same way as Instant-NGP\cite{muller2022instant_ngp}. 
%
Additionally, to achieve faster convergence, we sample 90\% rays on the foreground pixels and 10\% rays on the background pixels. 

%
%
\textbf{Hyperparameters.}
For static scene reconstruction, we train our model for 15k iterations, which takes around 5 minutes.
For dynamic scene reconstruction, we train the first frame from scratch for 2k iterations (4k iterations in particular for the Lego sequence due to its complex geometry), which takes about 40 seconds; then train each subsequent frame for 1.1k iterations, where we optimize the global transformation independently for the first 100 iterations and we fine-tune the network parameters and global deformation together for the remaining iterations.

%% file: supp_fig/0_fig_network_arch.tex
%
%
\begin{figure*}
    \centering
	\includegraphics[width=0.8\linewidth]{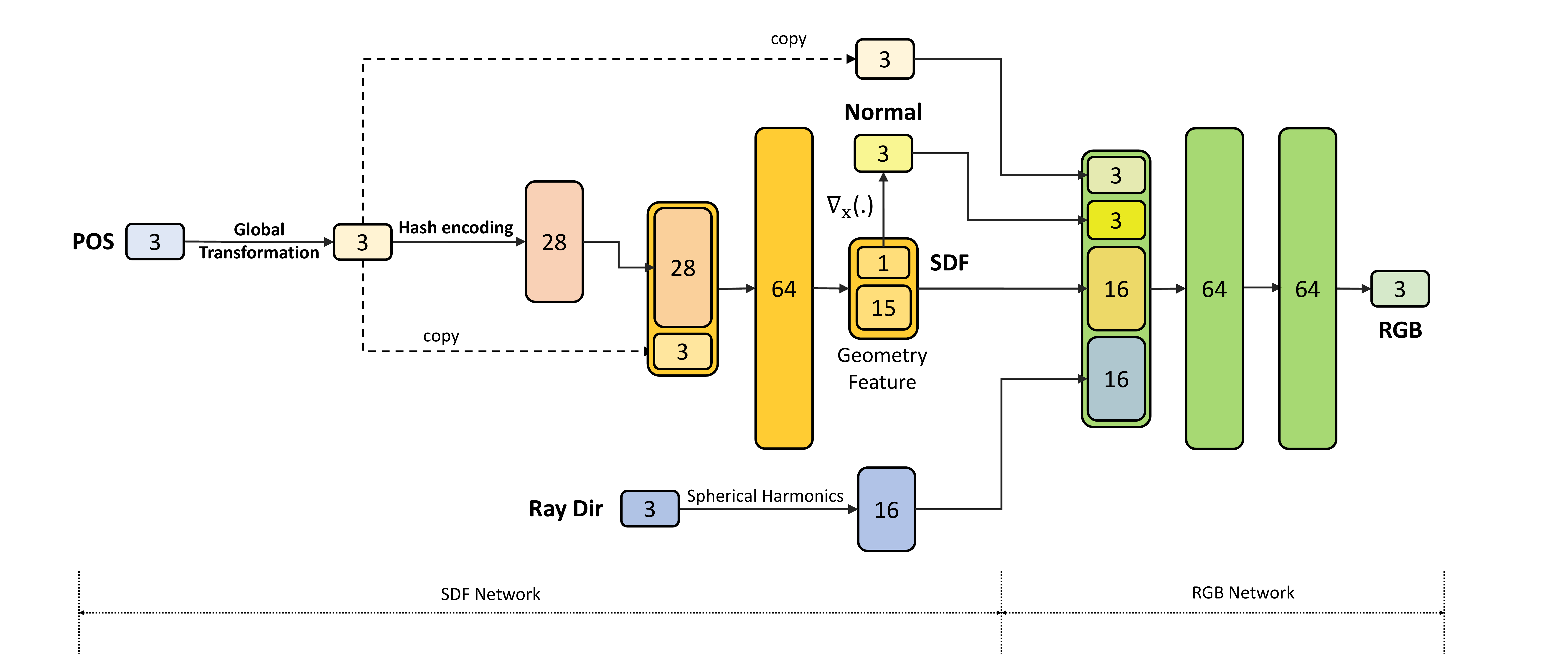}
	\vspace{-8pt}
	\caption
	{
	    A visualization of the network architecture of \textit{NeuS2}.
	}
	\label{fig:network_arch}
\end{figure*}
%
%